\documentclass[sigplan,nonacm]{acmart}
\usepackage{xcolor}
\usepackage{xspace}
\usepackage{mathtools} % upgrade from amsmath
\usepackage{bm}
\usepackage{enumitem}
\usepackage{textcomp}
\usepackage{changepage}
\usepackage{balance}
\usepackage{multicol}
\usepackage{array}

\usepackage{graphicx}
\usepackage{subcaption}
\usepackage{tikz}
\usepackage{tabularx}
\usepackage{multirow}
\usepackage{booktabs}
\usepackage{makecell}
\usepackage[figuresleft]{rotating}
\usepackage[flushleft]{threeparttable}
\usepackage{algorithmicx}
\usepackage{algorithm}
\usepackage[noend]{algpseudocode}

\algnewcommand\algorithmicinput{\textbf{Input:}}
\algnewcommand\Input{\item[\algorithmicinput]}
\algnewcommand\algorithmicoutput{\textbf{Output:}}
\algnewcommand\Output{\item[\algorithmicoutput]}

\usepackage[utf8]{inputenc} % Avoid insert Chinese characters
\usepackage[most]{tcolorbox} % Colored and framed text boxes
\usepackage{soul} % Colored and framed text
\usepackage{siunitx} % For \priority command
\usepackage{pifont} % inconsolata
\usepackage{xurl} % url break line
\usepackage{bbding} % icons
\usepackage[textsize=scriptsize, linecolor=magenta, bordercolor=magenta,
    backgroundcolor=white, textwidth=45pt]{todonotes}
\usepackage{marginnote}
\usepackage[noindentafter]{titlesec}
\titlespacing\section{0pt}{3pt}{3pt} %{left}{before}{after}
\titlespacing\subsection{0pt}{3pt}{3pt}
\titlespacing\subsubsection{0pt}{5pt}{5pt}

\usepackage[skip=3pt]{caption}
\newcommand{\SysName}{\textsc{SchedMate}\xspace}
\newcommand{\SysNameRM}{\textrm{SchedMate}\xspace}

\newcommand{\cmmnt}[1]{}

\begin{document}
\date{}

\newcommand{\papername}{Semantic-Aware Scheduling for GPU Clusters with Large Language Models}

\title{\Large \bf \papername}

\author{Zerui Wang}
\authornote{Both authors contributed equally to this research.}
\affiliation{%
  \institution{Shanghai Jiao Tong University \& Shanghai AI Laboratory}
  \city{Shanghai}
  \country{China}
}
\email{wangzerui@sjtu.edu.cn}

\author{Qinghao Hu}
\authornotemark[1]
\affiliation{%
  \institution{Nanyang Technological University}
  \city{Singapore}
  \country{Singapore}
}
\email{qinghao.hu@ntu.edu.sg}

\author{Ana Klimovic}
\affiliation{%
  \institution{ETH Zurich}
  \city{Zurich}
  \country{Switzerland}
}
\email{aklimovic@ethz.ch}

\author{Tianwei Zhang}
\affiliation{%
  \institution{Nanyang Technological University}
  \city{Singapore}
  \country{Singapore}
}
\email{tianwei.zhang@ntu.edu.sg}

\author{Yonggang Wen}
\affiliation{%
  \institution{Nanyang Technological University}
  \city{Singapore}
  \country{Singapore}
}
\email{ygwen@ntu.edu.sg}

\author{Peng Sun}
\affiliation{%
  \institution{Shanghai AI Laboratory}
  \city{Shanghai}
  \country{China}
}
\email{sunpeng@pjlab.org.cn}

\author{Dahua Lin}
\affiliation{%
  \institution{The Chinese University of Hong Kong}
  \city{Hong Kong}
  \country{China}
}
\email{dhlin@ie.cuhk.edu.hk}

\begin{abstract}
    Deep learning (DL) schedulers are pivotal in optimizing resource allocation in GPU clusters, but operate with a critical limitation: they are largely blind to the semantic context of the jobs they manage. This forces them to rely on limited metadata, leading to high profiling overhead, unreliable duration estimation, inadequate failure handling, and poor observability. To this end, we propose \SysName, a framework that bridges this semantic gap by systematically extracting deep insights from overlooked, unstructured data sources: source code, runtime logs, and historical jobs. \SysName enhances existing schedulers non-intrusively through three LLM-based components. Our implementation integrates seamlessly with existing deep learning schedulers. Evaluations on a 128-GPU physical cluster and extensive simulations on production traces show \SysName reduces average job completion times by up to 1.91$\times$, substantially enhancing the scheduling performance, demonstrating the critical role of semantic-awareness in modern DL scheduling.
\end{abstract}
\maketitle
\pagestyle{plain}

%-------------------------------------------------------------------------------
\section{Introduction}
\label{sec_intro}
%-------------------------------------------------------------------------------

Deep Learning (DL) has experienced unprecedented growth in recent years, with large language models (LLMs) like GPT-5~\cite{GPT-5}, LLaMA-3~\cite{llama3}, and GLM-4.5~\cite{GLM-4.5} pushing the boundaries of AI capabilities. This growth drives a substantial demand for computational resources, particularly multi-billion dollar GPU clusters~\cite{Philly,Helios,Acme} and tailored frameworks~\cite{HybridFlow,MegatronLM,InternEvo,Galvatron,Alpa}. DL schedulers are critical in orchestrating the execution of diverse and resource-intensive workloads within these clusters.

Despite extensive research on optimizing job completion time (JCT) and resource utilization~\cite{Pollux,Tiresias,Sia,Gavel,Lucid,AutoSched,Helios,Acme,AntMan,Horus,Prism,ONES}, existing schedulers are hampered by a fundamental \textbf{semantic gap}. They operate on surface-level metadata, leading to four significant real-world challenges:

\begin{figure}[t]
    \centering
    \includegraphics[width=1.0\linewidth]{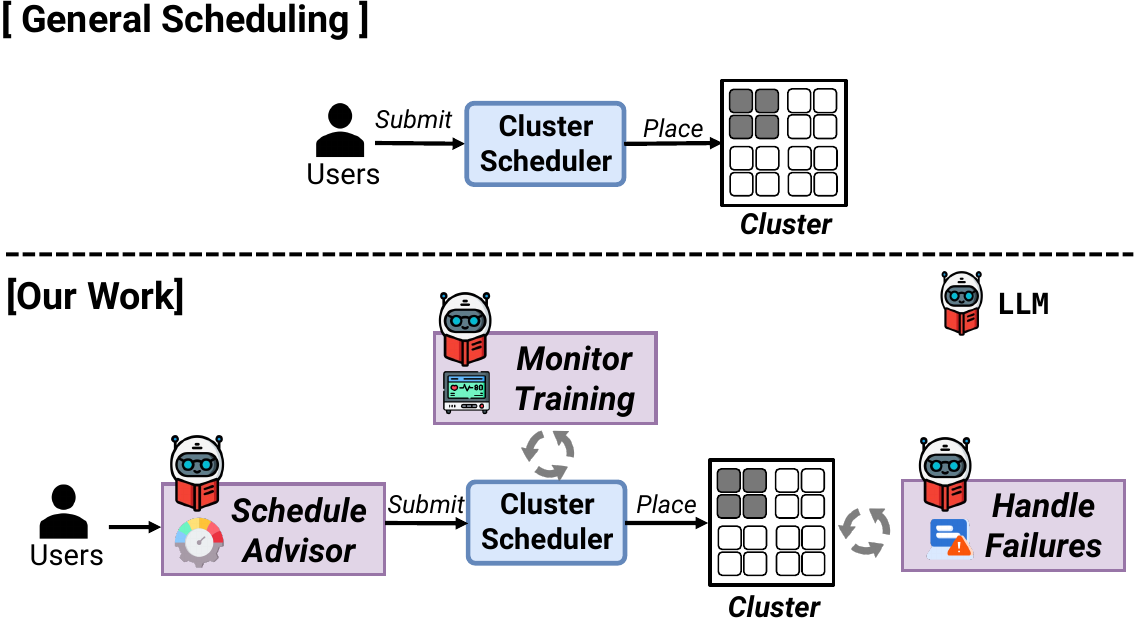}
    \caption{Existing DL schedulers operate on limited metadata (top), while \SysName enriches the scheduling process with deep insights from source code, logs, and historical jobs (bottom).}
    \label{figure_paradigm}
\end{figure}

\begin{figure}[t]
    \centering
    \includegraphics[width=1.0\linewidth]{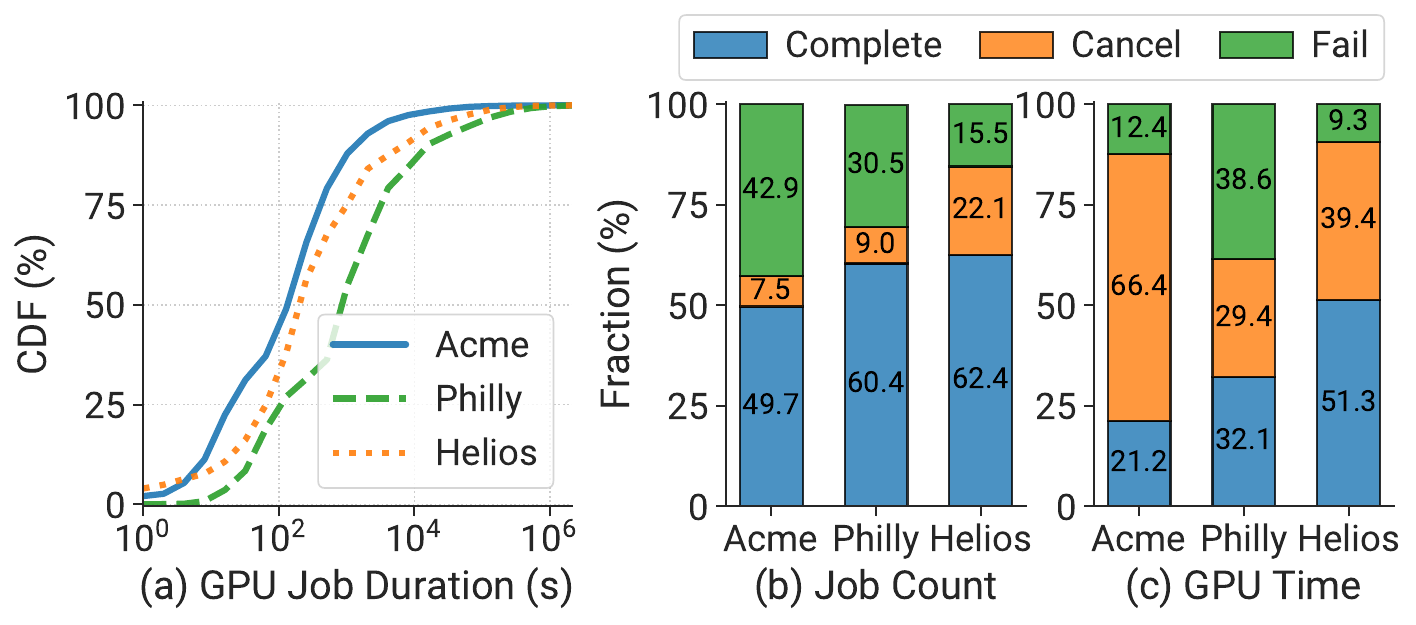}
    \caption{\textbf{Background}: DL workload characteristics across Microsoft Philly~\cite{Philly}, SenseTime Helios~\cite{Helios} and Acme~\cite{Acme} clusters. (a) CDF of the job duration. (b, c) Final statuses of jobs in terms of quantity and utilized GPU resources.}
    \label{fig:motivation_bar_cdf}
\end{figure}

\begin{itemize}[leftmargin=*,topsep=-1pt, itemsep=0pt, itemindent=8pt]
    \item \textbf{C1: High Profiling Overhead.} To forecast resource needs, many schedulers~\cite{Sia,Gavel,Lucid,Themis} rely on job profiling. This process is prohibitively expensive, especially since most production jobs are short-lived. For instance, the median job duration in the Acme cluster is just two minutes~\cite{Acme} (Figure~\ref{fig:motivation_bar_cdf}a), meaning profiling can consume as much time as the job's entire execution. For distributed LLM training, this overhead can occupy hundreds of GPUs, imposing a significant operational burden on the cluster.

    \item \textbf{C2: Unreliable Duration Estimation.} Duration-aware policies~\cite{Themis,Gavel,Lucid,MLaaS,Helios} depend on accurate job runtime predictions, but current methods can be unreliable in production scenarios. Intrusive techniques~\cite{Themis,Pollux,Singularity,AFS,Gandiva}  that require users to provide \texttt{num\_steps} and \texttt{step\_time} fail because they assume jobs run to completion. However, production data shows that up to half of all jobs terminate prematurely~\cite{Acme,Philly,Helios} (Figure~\ref{fig:motivation_bar_cdf}b), often rendering user-provided estimates inaccurate. ML-based methods~\cite{Lucid,MLaaS} typically shows poor prediction accuracy, as they use only sparse metadata (e.g., job name, username) and lack the rich features needed for precise predictions.

    \item \textbf{C3: Inadequate Failure Handling.} Job failures are frequent and costly in large-scale training~\cite{Acme,Philly,llama3,OPT}, yet most state-of-the-art schedulers~\cite{Gavel,Pollux,Sia,Lucid} lack robust failure-handling mechanisms. For example, the pretraining of LLaMA-3.1 encountered 419 failures, averaging one every three hours~\cite{llama3}. This oversight leads to wasted resources and significant delays, especially for large distributed jobs where rapid recovery is essential.

    \item \textbf{C4: Limited Observability.} Schedulers are often blind to runtime dynamics like training progress or performance regressions. This creates a dilemma: \textit{non-intrusive} schedulers are easy to deploy but see only surface-level metadata, while \textit{intrusive} schedulers that modify DL frameworks to gain deeper observability are brittle and impractical. They impose a high maintenance burden~\cite{Lucid} and cannot keep pace with the rapid evolution of frameworks~\cite{PyTorch,TensorFlow}. The intrusive scheduler Pollux, for example, is now incompatible with modern PyTorch due to years of inactivity~\cite{Pollux}. Consequently, most production clusters~\cite{Helios,Acme,Philly} opt for non-intrusive approaches, sacrificing observability.
\end{itemize}

% Key insight: source code and log contain information -> can be used to
These challenges stem from a common root cause: the scheduler's inability to access the rich \textit{semantic information} of a DL job. We argue that this critical information is not missing but is instead locked away in unstructured data artifacts that have been traditionally overlooked: the job's \textit{source code} and \textit{runtime logs}. Source code contains inherent details about workload characteristics (for \textbf{C1}, \textbf{C2}), while logs record runtime dynamics and failure messages (for \textbf{C3}, \textbf{C4}). The key research questions are \textbf{what semantic information to extract}, \textbf{how to extract it efficiently}, and \textbf{how to use it to enhance scheduling} without imposing new burdens on users or operators.

To bridge this semantic gap, we introduce \SysName, a framework that unlocks this information to enable a new paradigm of \textit{semantic-aware scheduling}. \SysName integrates seamlessly into existing scheduling workflows via three LLM-powered modules (Figure~\ref{figure_paradigm}): (1) The \textbf{Scheduling Advisor} addresses \textbf{C1} and \textbf{C2} by using an LLM agent to analyze job source code. It extracts key workload metadata to find similar historical jobs, enabling accurate workload prediction without costly profiling. (2) The \textbf{Metric Tracker} tackles \textbf{C4} with a non-intrusive, two-stage pipeline to extract training progress from verbose logs. By using efficient embedding-based filtering followed by precise LLM-based extraction, it provides real-time visibility for dynamic scheduling adjustments. (3) The \textbf{Failure Handler} resolves \textbf{C3} by performing automated root cause analysis on logs. It rapidly pinpoints error messages, identifies the cause, and can trigger automated recovery actions for infrastructure-related issues, minimizing downtime.

We implement \SysName as a plug-and-play module and demonstrate its effectiveness through three case studies: enhancing the non-intrusive scheduler Lucid~\cite{Lucid}, the elastic scheduler Sia~\cite{Sia}, and as a standalone scheduler. We conduct comprehensive experiments on a 128-GPU physical cluster and extensive simulations on public production traces, from Microsoft~\cite{Philly}, SenseTime~\cite{Helios}, and Acme~\cite{Acme}. To enable fine-grained analysis, we also introduce Mars, a new trace collected from our in-house cluster containing rich source code, logs, and metrics from real-world large-scale LLM development jobs. The results show that \SysName reduces average job completion time by up to 1.91$\times$. These findings demonstrate that by bridging the semantic gap, \SysName effectively addresses challenges \textbf{C1–C4} and significantly enhances the performance of modern DL schedulers.

%-------------------------------------------------------------------------------
\section{Background and Motivation}
\label{sec_background}

\subsection{DL Workload Scheduling}\label{subsec_backgorund}

\noindent\textbf{Existing DL Schedulers.}~DL jobs are usually submitted to multi-tenant GPU clusters~\cite{Acme,MLaaS,Helios,Philly}. Figure \ref{figure_paradigm} shows common paradigms of DL workload scheduling~\cite{SurveyDLSched,SLURM,MLaaS}. Schedulers manage these jobs based on policies like First-In-First-Out (FIFO) or Shortest Job First (SJF)~\cite{SurveyDLSched}. To optimize for goals like minimizing job completion time (JCT), advanced schedulers employ techniques such as \textit{job-packing} (co-locating jobs)~\cite{Lucid, AntMan,Prism}, \textit{elastic-scheduling}~\cite{Pollux,Sia,Singularity,ELAN}, and \textit{heterogeneous-aware policies}~\cite{Gavel,Gandivafair,Sia}. Some schedulers also focus on other specific like model-targeted scheduling~\cite{ModelKeeper,Serving22}, inference workloads~\cite{AlpaServe,DeepPlan,MIGServing}.

\noindent\textbf{Workload Estimation.}~Effective scheduling requires estimating workload characteristics, such as duration and resource utilization. Schedulers gather necessary metrics via pre-profiling (briefly running jobs beforehand, leveraging their iterative nature~\cite{Gavel,Lucid}) or online profiling (collecting metrics during execution~\cite{Pollux,Sia}). Two main duration estimation approaches are used: (1) \textit{Config-based} methods calculate $step\_time \times total\_steps$~\cite{Themis,Gavel,optimus,PCS,Pollux}, but are unreliable under high failure/cancellation rates. (2) \textit{History-based} methods exploit recurring job submission patterns~\cite{Helios,MLaaS,Acme}. They usually utilize the resource profile and scheduling information of historical jobs to train predictive models~\cite{MLaaS,Helios,Lucid} or directly identify similar historical jobs~\cite{3Sigma,Slearn} for workload estimation.

\noindent\textbf{Prevalence of Failures.}~Recent studies~\cite{Philly,Acme,llama3,jiang2024MegaScaleScaling} report that DL training jobs suffer from frequent failures, especially infrastructure failures. Acme \cite{Acme} reports around 2 infrastructure failures per day on average during LLM development. ByteDance~\cite{jiang2024MegaScaleScaling} experienced over 100 failures during a multi-week LLM pretraining task. As shown in Figure \ref{fig:motivation_bar_cdf}(c), it is obvious that only approximately 20$\sim$50\% of resources are consumed by jobs that are finally complete, demonstrating the importance of failure handling.

Despite their sophistication, these schedulers operate on a thin layer of structured metadata, leaving them blind to the rich, semantic context of the workloads they manage, leading to the challenges outlined in the \S\ref{sec_intro}.

\subsection{The Opportunity: Semantic-Aware Scheduling}

We argue that the critical information needed to bridge this semantic gap already exists within the cluster but is locked away in unstructured data artifacts traditionally ignored by schedulers: a job's \textbf{source code}, its \textbf{runtime logs}, and the collective history of \textbf{all jobs}. Production traces confirm these artifacts are widely available. For instance, production GPU clusters of Microsoft~\cite{Philly}, SenseTime~\cite{Helios}, and Acme~\cite{Acme} have access to the complete source code and logs for every job.  The core challenge of this paper is how to systematically extract and leverage this semantic information for enhancing the scheduling of state-of-the-art cluster schedulers.

\begin{figure}[t]
    \centering
    \includegraphics[width=1.0\linewidth]{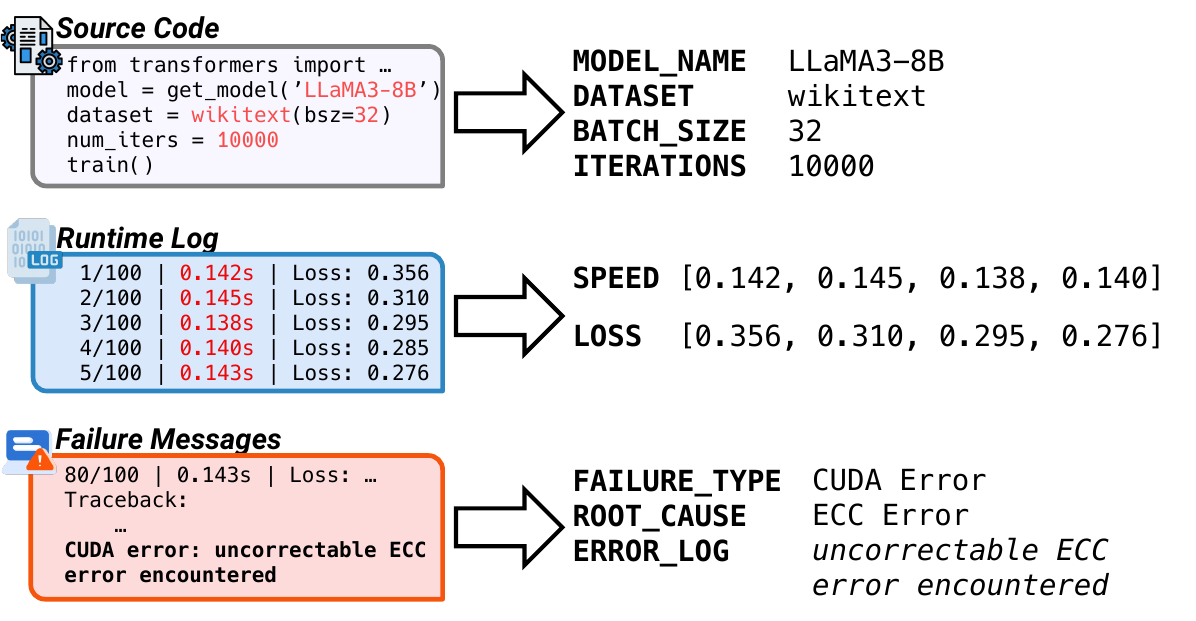}
    \caption{Examples of semantic information available in source code and runtime logs that are opaque to traditional schedulers.}
    \label{fig:extraction_example}
\end{figure}

\noindent\textbf{Why Large Language Models?}~The information in source code and logs is unstructured and context-dependent. Traditional tools are too brittle to parse this data reliably across different DL frameworks and user code. LLMs are well-suited for this diverse task, as they can understand natural language and code structure, allowing them to perform the reasoning needed to extract workload metadata, find metrics in noisy logs, and diagnose failures. Although many recent works~\cite{LogPPT,LLMParser} have studied how LLMs can be used to skillfully perform similar information extraction tasks, to the best of our knowledge, we are the first to harness LLM-extracted information to enhance the performance of DL schedulers.

\noindent\textbf{Semantic Workload Similarity.}~Source code contains information of workload characteristics, available at submission time, but \textit{how to leverage it for workload prediction, especially resource profile and duration prediction?} We find that jobs with similar workloads often exhibit similar resource profiles and duration distributions. For example, Acme~\cite{Acme} shows that jobs with different workload types (MLLM, pretrain, SFT) exhibit distinct duration distributions, a pattern we also observe in our Mars trace. This suggests that we can use semantic information from source code to \textit{identify similar historical jobs}, which can then be used to predict the resource profile and duration of new jobs. Existing history-based schedulers define similarity by matching basic metadata~\cite{MLaaS,Lucid}. We demonstrate that a better workload similarity is semantic, which addresses challenges \textbf{C1} and \textbf{C2}.

\noindent\textbf{Observability from Logs.}~Runtime logs offer a non-intrusive window into job progress through metrics like \texttt{step\_time} and \texttt{loss}, which is ideal for production schedulers that prefer non-intrusive schedulers like Lucid~\cite{Lucid}. It is straightforward to use LLM to non-intrusively parse logs and extract these metrics, but concerns are that logs are often verbose and may not contain the necessary information. Firstly, we can remove irrelevant logs by embedding-based filtering before LLM processing, which incurs minimal overhead compared to LLMs~\cite{RCACopilot}. Furthermore, considering diverse scenarios, we can develop a DL scheduler enhancement that works when only part of the logs provides the needed information and falls back to the default behavior when the logs are insufficient. This addresses challenge \textbf{C4}.

\noindent\textbf{Failure Handling with Logs.}~Production data shows a key distinction among job failures. Application-level bugs (e.g., syntax errors) require manual intervention. In contrast, \textit{infrastructure failures} (e.g., network issues, hardware faults) are often recoverable by restarting the job~\cite{Acme,Philly}. As shown in Figure~\ref{fig:motivation_bar_cdf}c failed jobs consume significant GPU resources. Automatically handling these failures can significantly reduce resource waste and enhance scheduling performance. Existing schedulers do not handle failures well. We are motivated to study how failure handling enhances DL scheduling, thereby addressing challenge \textbf{C3}.

In summary, our approach is not merely about applying LLMs to logs and code, which is a well-studied area~\cite{LogPPT,LLMParser}. Our core contribution is twofold: (1) identifying what specific semantic information is crucial for scheduling (workload similarity, runtime progress, failure type), and (2) designing mechanisms to leverage this information to facilitate semantic-aware scheduling.
\begin{figure}[t]
  \centering
  \includegraphics[width=1.0\linewidth]{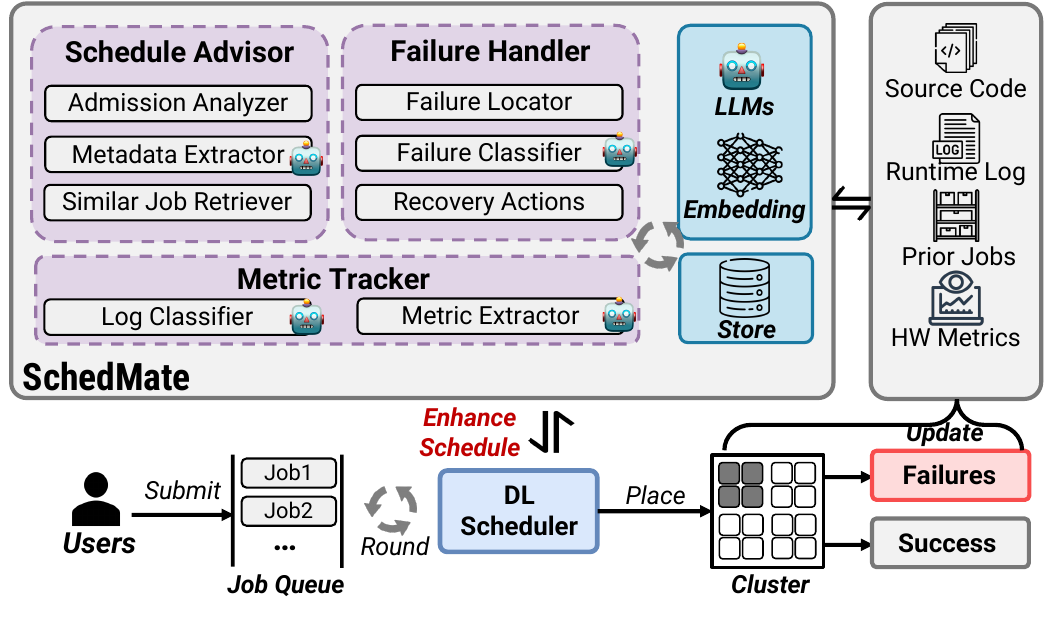}
  \caption{\textbf{System Overview of \SysName} The system integrates with four data sources (prior jobs, source code, runtime log, and hardware metrics). The modules attached to robot symbols utilize LLMs. \cmmnt{The system integrates with four data sources (prior jobs, source code, runtime log, and hardware metrics). Main components: (1) the \textit{Scheduling Advisor}, which performs similar job retrieval and workload estimation before scheduling; (2) the \textit{Metric Tracker}, which enables progress-aware decision making during training by analyzing runtime logs; and (3) the \textit{Failure Handler}, which manages failure parsing and recovery actions in the post-training phase.}}
  \label{figure_system}
\end{figure}

\section{\SysNameRM Overview and Design}
\label{subsec_system_overview}

\noindent\textbf{System Overview.} Figure \ref{figure_system} presents the architecture of \SysName, a framework designed to enable semantic-aware scheduling. It bridges the semantic gap of DL schedulers by extracting and integrating semantic information from four key data sources: source code, runtime logs, historical job data, and hardware metrics. \SysName consists of three core modules: the \textit{Scheduling Advisor}, the \textit{Metric Tracker}, and the \textit{Failure Handler}. The \textit{Scheduling Advisor} analyzes source code to extract semantic workload characteristics, enabling accurate predictions without profiling. The \textit{Metric Tracker} non-intrusively parses runtime logs to extract performance metrics, providing real-time observability into job progress. The \textit{Failure Handler} analyzes logs to semantically diagnose job failures and automate recovery.

\noindent\textbf{Semantic Workload Prediction.} To replace costly profiling and improve workload estimation, the \textit{Scheduling Advisor} uses a retrieval-based approach. The core idea is to leverage semantic information from a new job's source code to find similar historical jobs. The performance data from these past jobs is then used to predict the new job's characteristics, such as its duration and resource utilization. To efficiently extract this semantic metadata from code, we developed an LLM-based agent that reasons about the code structure to summarize key workload characteristics (in \S\ref{subsec_scheduling_advisor}).

\noindent\textbf{Progress-aware Decision Making.} The \textit{Metric Tracker} provides schedulers with real-time information of a job's runtime progress by extracting performance metrics from logs. To handle verbose and varied log formats efficiently, it employs a two-stage pipeline to efficiently filter irrelevant lines and extract structured metrics. This grants schedulers the observability needed for dynamic, progress-aware optimizations. The component's design is discussed in \S\ref{subsec_metric_tracker}.

\noindent\textbf{LLM-based RCA \& Failure Handling.} Infrastructure failures are a major source of wasted resources, yet schedulers lack the semantic context to handle them intelligently. Our \textit{Failure Handler} provides this context through automated root cause analysis (RCA). It uses an efficient method to first locate the initial error in massive logs, then employs an LLM to classify the failure's semantic type (e.g., infrastructure vs. application). For infrastructure failures, the system can trigger automated recovery actions, minimizing manual intervention and cluster-wide resource idleness (\S\ref{subsec_failure_handler}).

\begin{figure}[t]
  \centering
  \includegraphics[width=0.80\linewidth]{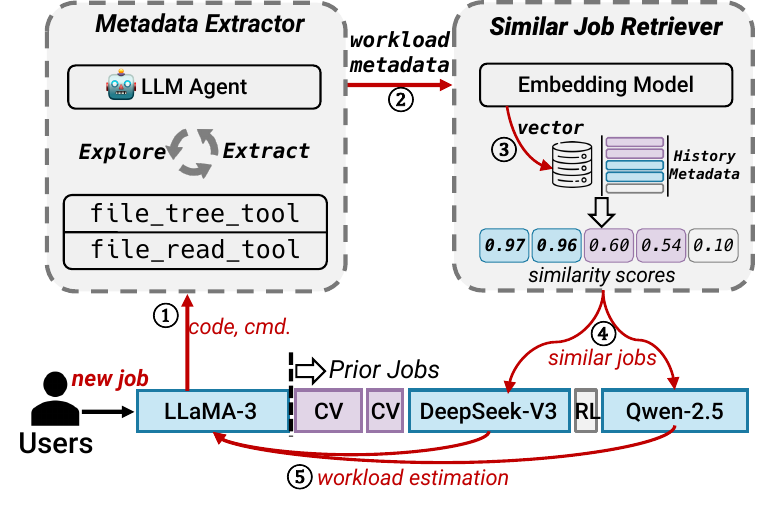}
  \caption{\textbf{Scheduling Advisor Workflow.} The LLM-based agent leverages two tools to support the efficient metadata extraction from source code. Length of each job refers to the duration. The red line indicates the data flow path.}
  \label{fig:scheduling_advisor}
\end{figure}

\subsection{Scheduling Advisor}\label{subsec_scheduling_advisor}

The \textit{Scheduling Advisor} provides workload insights by retrieving similar historical jobs, avoiding intensive profiling. It employs a retrieval-based approach that extracts metadata from a job and identifies similar historical jobs based on metadata similarity.

\noindent\textbf{Semantic Workload Metadata.} To capture the features of a workload, we focus on extracting key semantic metadata from the source code. Since source code can be extensive, we target the most informative details: (1) \textit{Model Arch}, including model name, type, and task (e.g., NLP, CV); (2) \textit{Dataset Settings}, such as dataset name; and (3) \textit{Training Configuration}, like training steps, and hardware requirements.

\noindent\textbf{Workflow.} As shown in Figure \ref{fig:scheduling_advisor}, when a job is submitted, the \textit{Metadata Extractor} analyzes its source code to extract this semantic metadata. The \textit{Similar Job Retriever} then uses this metadata to find matching historical jobs. Finally, the historical performance data of these similar jobs is aggregated to produce a workload estimation for the new job, which is then used by the scheduler.

\begin{figure}[t]
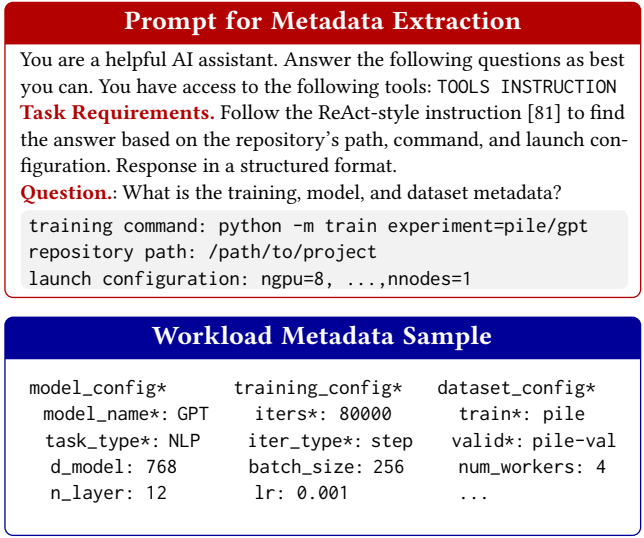

    \centering
    % \vspace{-15pt}
    % \begin{tcolorbox}[colback=white,colframe=gray!50!black,left=1mm, right=1mm, top=0mm, bottom=0mm, arc=1mm, boxrule=0.5pt, fontupper=\footnotesize]
    %     Log
    % \end{tcolorbox}

    \begin{tcolorbox}[colback=white,colframe=red!70!black, left=1mm, right=1mm, top=0mm, bottom=0mm, arc=1mm, boxrule=0.5pt, halign title={center}, fontupper=\footnotesize, fontlower=\footnotesize, title=\textbf{Prompt for Metadata Extraction}]
  You are a helpful AI assistant. Answer the following questions as best you can. You have access to the following tools:  \texttt{TOOLS INSTRUCTION}\\
          % \textcolor{red!70!black}{\textbf{ReAct Instruction.}}
          % Use the following format:\\
          % Question: the input question you must answer \\
          % Thought: you should always think about what to do \\
          % Action: the action to take, should be one of the tools: [file reader, file tree reader] \\
          % Action Input: the input to the action\\
          % Observation: the result of the action\\
          % ... (this Thought/Action/Action Input/Observation can repeat N times)\\
          % Thought: I now know the final answer\\
          % Final Answer: the final answer to the original input question.\\
          \textcolor{red!70!black}{\textbf{Task Requirements.}}
          Follow the ReAct-style instruction~\cite{yao2023react} to find the answer based on the repository's path, command, and launch configuration. Response in a structured format.\\
          \textcolor{red!70!black}{\textbf{Question.}}: What is the training, model, and dataset metadata?

          \tcbox[on line, colframe=white, boxrule=0.1pt, colback=gray!10!white, width=100pt, height=30pt, top=0pt,left=0pt,right=0pt,bottom=0pt]{
                \begin{minipage}[t]{224pt}
                      \texttt{training command: \texttt{python -m train experiment=pile/gpt}\\
                      repository path: \texttt{/path/to/project}\\
                      launch configuration: \texttt{ngpu=8, ...,nnodes=1}}
                \end{minipage}
          }
    \end{tcolorbox}

      \vspace{-5pt}
      \begin{tcolorbox}[colback=white,colframe=blue!60!black, left=1mm, right=1mm, top=0mm, bottom=-1mm, arc=1mm, boxrule=0.5pt, halign title={center}, fontupper=\footnotesize, title=\textbf{Workload Metadata Sample}]
        \begin{tcbraster}[raster columns=3, raster equal height, raster column skip=1mm]
            \begin{tcolorbox}[colback=white, left=0mm, right=1mm, top=0.5mm, bottom=0mm, arc=1mm, boxrule=0.5pt, enhanced, frame hidden]
        \begin{verbatim}
model_config*
  model_name*: GPT
  task_type*: NLP
  d_model: 768
  n_layer: 12
      \end{verbatim}
            \end{tcolorbox}
            \begin{tcolorbox}[colback=white, left=0mm, right=1mm, top=0.5mm, bottom=0mm, arc=1mm, boxrule=0.5pt, enhanced, frame hidden]
        \begin{verbatim}
training_config*
  iters*: 80000
  iter_type*: step
  batch_size: 256
  lr: 0.001
      \end{verbatim}
            \end{tcolorbox}
            \begin{tcolorbox}[colback=white, left=0mm, right=1mm, top=0.5mm, bottom=0mm, arc=1mm, boxrule=0.5pt, enhanced, frame hidden]
            \begin{verbatim}
dataset_config*
  train*: pile
  valid*: pile-val
  num_workers: 4
  ...
      \end{verbatim}
            \end{tcolorbox}
            \end{tcbraster}
          \end{tcolorbox}
    \caption{Prompt template of \textit{Metadata Extractor} and a workload metadata sample. Fields with * are mandatory.}
    \label{figure_llm_prompt}
  \end{figure}

\noindent\textbf{Metadata Extractor.} This component is designed to extract semantic workload metadata from the job's source code. This information is often scattered across multiple files, making simple rule-based extraction brittle. To address this, we use a tool-based ReAct-style agent~\cite{yao2023react} that intelligently navigates the codebase. We equip the agent with two simple filesystem tools: \texttt{file\_tree\_tool}, which provides a filtered view of the project structure, and \texttt{file\_read\_tool}, which allows the agent to inspect file contents.

The agent's prompt is shown in Figure \ref*{figure_llm_prompt}. The prompt provides the agent with: (1) the task definition, (2) instructions for using the tools (the ReAct framework), (3) the desired metadata schema, and (4) formatting instructions to ensure the output is a structured JSON. This setup enables the agent to autonomously explore the codebase, identify relevant files (like configuration scripts or model definitions), and extract the specified semantic metadata efficiently.

The agent's process, illustrated in Figure~\ref{fig:scheduling_advisor}, is iterative. It starts with the job's working directory and command, uses the tools to explore and read files, and reasons about the gathered information to decide its next action. This continues until it has collected sufficient metadata. The final output is a structured JSON summary, providing a comprehensive semantic fingerprint of the job's workload for the retriever.

\begin{figure}[t]
  \centering
  \includegraphics*[width=0.85\linewidth]{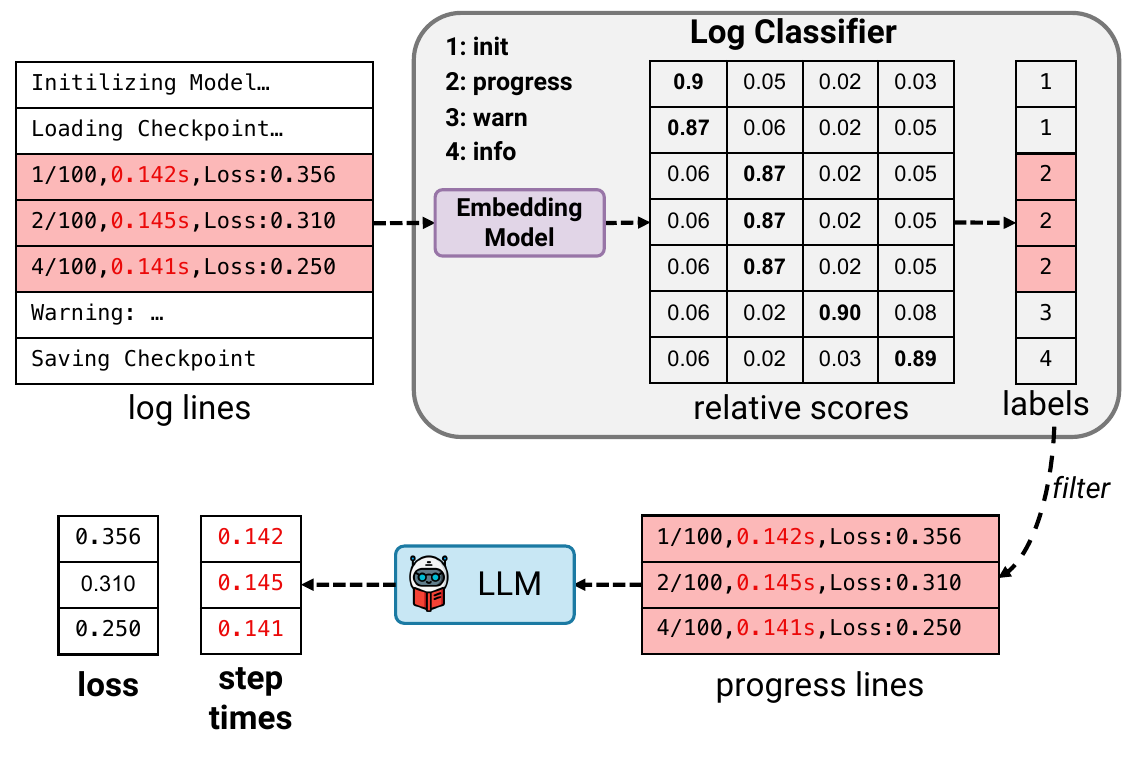}
  \caption{\textbf{Metric Tracker Workflow}. The progress lines are blended with some initialization and warning logs.}
  \label{fig:metric_tracker}
\end{figure}

\noindent\textbf{Similar Job Retrieval.}~To match the extracted semantic metadata with historical jobs, we use a similarity-based retrieval approach. Although the metadata has fixed fields, we still cannot directly use them for scheduling, because the values are dynamic. Therefore, instead of directly using the value of metadata for text-level operation, we use an embedding-based method. Specifically, the structured metadata is first embedded into a dense vector using an embedding model (e.g. BGE-m3~\cite{bge2024chen}). We then compute the \textit{cosine similarity} between the new job's vector and the vectors of all historical jobs. Historical jobs with a similarity score above a threshold (T
score) are considered matches. We then select the \textit{top-k} most similar jobs to inform the workload estimation step. This semantic matching process effectively identifies jobs with truly similar workload characteristics.

\noindent\textbf{Workload Estimation from Similar Jobs.} Once the top-k similar historical jobs are identified, their performance data is used to predict the new job's characteristics. Specifically, we calculate the average duration and average resource utilization (e.g., SM utilization) of these similar jobs. This average value serves as the final estimation for the new job, providing a reliable forecast rooted in the performance of semantically similar past workloads. This estimation is then passed to the scheduler to inform its decisions, such as job ordering or packing. We also showcase another method of using similar jobs for performance modeling in \S\ref{subsec_sia_case_study}.

\subsection{Metric Tracker}\label{subsec_metric_tracker}

The \textit{Metric Tracker} offers runtime observability, providing schedulers with real-time insights into job performance without intrusive code modifications. But this information is buried within thousands of unstructured log lines. Directly applying LLMs to parse entire log streams is computationally prohibitive and introduces unacceptable latency for real-time scheduling. To solve this, we designed a lightweight, two-stage pipeline: a fast classifier first filters for relevant lines, and a more powerful LLM then parses only this small subset. This design achieves high-fidelity metric extraction with the low latency required for dynamic scheduling decisions. The workflow is shown in Figure \ref{fig:metric_tracker}.

\noindent\textbf{Log Classifier.} Production DL logs are highly heterogeneous~\cite{Acme}. To handle this diversity, the \textit{Log Classifier} first categorizes each log line into high-level semantic types, as shown in Figure~\ref{fig:metric_tracker}. The classification is performed by computing the cosine similarity between a line's embedding vector and category vectors. These category vectors are generated once by embedding descriptive text labels and generalize well across different logging styles. This step acts as a fast, semantic filter, isolating the small subset of lines likely to contain progress metrics.

\noindent\textbf{Metric Extractor.} Once the \textit{Log Classifier} identifies progress-related lines, the \textit{Metric Extractor} uses an LLM to parse them and extract structured data (e.g., step time, loss). To ensure low latency, we process logs in reverse chronological order and stop after successfully extracting metrics from a small number (N) of lines, which is typically sufficient for a stable estimate. To ensure robustness, we also further remove outliers. This two-stage design is highly cost-effective, as it leverages the much higher throughput of embedding models for the initial bulk filtering, reserving the more powerful but slower LLM for the final, precise extraction task.

\begin{figure}[t]
  \centering
  \includegraphics[width=1.0\linewidth]{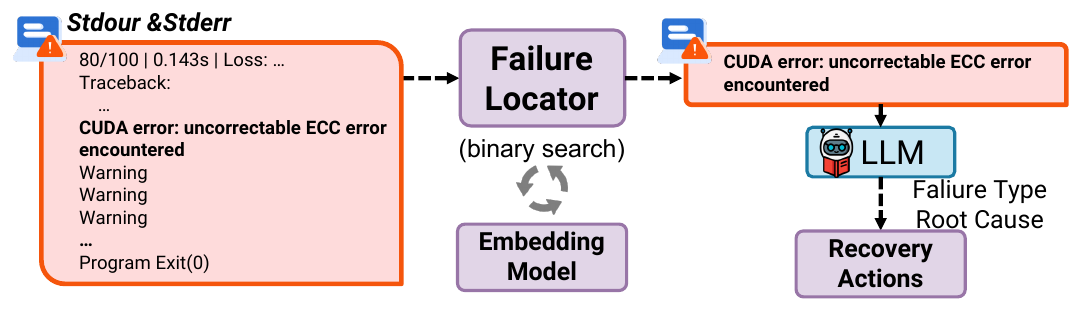}
  \caption{\textbf{Failure Handler Workflow}. The emphasized log line is the first failure we aims to locate.}
  \label{fig:failure_handler}
\end{figure}

\noindent\textbf{Robustness to Partial Information.}~The effectiveness of the \textit{Metric Tracker} depends on jobs logging their training progress. If a job does not output this information, the tracker cannot provide metrics. Our integrations into existing schedulers are designed to be robust to this scenario: if needed metrics cannot be extracted for a particular job, the scheduler falls back to its default behavior for that job, ensuring no degradation in scheduling performance.

\subsection{Failure Handler}\label{subsec_failure_handler}

The \textit{Failure Handler} is designed to close the semantic gap in failure management, a major source of wasted resources in GPU clusters (\S\ref{sec_background}). Its goal is to empower the scheduler to distinguish between recoverable \textit{infrastructure failures} (e.g., faulty hardware) and non-recoverable \textit{application-level errors} (e.g., code bugs). The core technical challenge is locating the single root-cause message within massive logs that are often filled with thousands of cascading, secondary error messages. To solve this efficiently, the \textit{Failure Handler} employs a three-stage pipeline: fast failure message localization, LLM-based classification, and automated recovery.

\noindent\textbf{Failure Locator.} To efficiently find the root cause in a haystack of error messages, the \textit{Failure Locator} uses a binary search algorithm guided by our semantic \textit{Log Classifier}. This approach is based on the empirical observation that the first true error message typically triggers a cascade of subsequent errors. This creates a semi-sorted log structure (normal logs → error logs), which is ideal for binary search. The algorithm repeatedly splits the log and uses the \textit{Log Classifier} to check the midpoint. To improve robustness against normal logs that may be interleaved with errors after a fault, the search operates on log \textit{chunks} rather than single lines. A chunk is classified as containing an "error" if it includes any failure-related messages, allowing the search to reliably converge on the initial failure.

\noindent\textbf{Failure Classification.} Once the initial error message is located, a small context window (e.g., 500 lines) around that point is passed to an LLM for fine-grained classification. The failure is categorized based on a taxonomy adapted from production cluster analysis~\cite{Acme}, distinguishing between infrastructure failures, framework errors, and user script bugs. For infrastructure failures, the LLM performs a second classification step to identify the likely faulty component (e.g., GPU, NVLink, Node). To ensure structured output with two fields: \textit{Error Type} and \textit{Faulty Component}.

\noindent\textbf{Automated Recovery Actions.} Once the failure is classified as infrastructure-related and a faulty component is identified, \SysName autonomously executes a predefined recovery action. For example, if a GPU failure is diagnosed, the system can: (1) run a diagnostic tool like \texttt{nccltest} to confirm the fault, (2) isolate the node in the cluster manager to prevent future placements, (3) provision a replacement node, and (4) restart the job from its last checkpoint on the new resources. This automated remediation pipeline eliminates the need for manual intervention for a common and costly class of failures, significantly reducing job downtime.

\section{Integration Case Studies}\label{sec_cases}

\subsection{Case Study 1: Non-intrusive Scheduler Lucid}
\label{subsec_integration_nonintrusive}

Lucid~\cite{Lucid} is a non-intrusive scheduler that uses job packing to minimize Job Completion Times (JCTs), guided by metrics from a brief, initial profiling period ($T_{prof}$). While effective, this design introduces several challenges in production environments.

In practice, however, this design faces several challenges that limit its effectiveness. First, the brief profiling can be inaccurate. If $T_{prof}$ occurs during a job's initial "warmup" stage, it will underestimate the job's true steady-state resource utilization~\cite{CharacCGS}. This frequently leads to suboptimal packing decisions, such as co-locating two resource-intensive jobs, which causes significant interference and performance degradation. Second, Lucid's duration estimator relies on limited, surface-level metadata (e.g., username, job name), which is insufficient for capturing the true workload characteristics of a job. Finally, Lucid lacks any mechanism to detect or react to these packing-induced slowdowns at runtime, meaning poor decisions cannot be corrected.

We integrate \SysName to address these specific limitations with three enhancements:

\noindent\textbf{Bypass Profiling.} To overcome inaccurate profiling, we use the \textit{Scheduling Advisor}. For a new job, the Advisor retrieves similar historical jobs. If a match is found ($k>0$), we use the average of their historical hardware metrics as a more reliable prediction, bypassing the default profiler. If no match is found ($k=0$), the system falls back to default profiling.

\noindent\textbf{Enhanced Duration Estimation.} We replace Lucid's limited model with the \textit{Scheduling Advisor}'s semantic, retrieval-based approach. We use the average duration of the matched similar jobs as the new prediction. If no match is found, we fall back to Lucid's original estimator.

% Figure~\ref{fig:duration_estimation_error_lucid} shows that the \textit{Scheduling Advisor} achieves higher accuracy compared to Lucid's workload estimator. Specifically, our method maintains an estimation error within 100\%.0 for 85\% of the cases, whereas Lucid's estimator only achieves 21\%.

\noindent\textbf{Interference-aware Packing Cancelation.}~To provide the missing feedback loop of over-slowdown, we introduce an non-intrusive interference-aware packing cancelation mechanism using the \textit{Metric Tracker}. We take two jobs, A and B, for example. Initially, only Job A is running in the cluster. The \textit{Metric Tracker} non-intrusively process Job A's log and obtain $TP_{before}$, the throughput of job A before packing is conducted. Then, Job B is scheduled and packed with Job A. After a period of time, which ensures passing the warmupstage, we then collect the newly generated log of Job A and get $TP_{after}$, the throughput after packing. We require that Job A has completed several tens of training steps after packing before the second triggering of \textit{Metric Tracker} to unsure steady running of the two packed jobs. Finally, if we detect that the slowdown rate ($\frac{TP_{after}}{TP_{before}}$) of Job A is lower than a threshold (35\% slowdown in the later experiments), the scheduler intelligently cancels the packing by evicting Job B and returns it back to the job queue for rescheduling. Considering that the model training progress of the evicted jobs will be lost, we restrict that the evicted jobs cannot be packed again to avoid frequent eviction. Note that the eviction check is skipped for the specific job pairs where the job's progress cannot be tracked.

\subsection{Case Study 2: Elastic Scheduler Sia}
\label{subsec_integration_elastic}

Sia~\cite{Sia} is a state-of-the-art heterogeneity-aware, elastic scheduler designed to improve JCTs and resource utilization in heterogeneous clusters. Sia begins each job by profiling it and then adopts a profile-as-you-go approach, using this information to make scheduling decisions. It records all throughput information under all placements and batch sizes encountered by each job. However, if the initial profiling process is conducted for every job, it can be both costly and time-consuming. To address this limitation, we have integrated \SysName into Sia, minimizing the profiling overhead by identifying resubmissions or similar jobs and bypassing their profiling process.

\begin{figure}[t]
    \centering
    \includegraphics[width=1.0\linewidth]{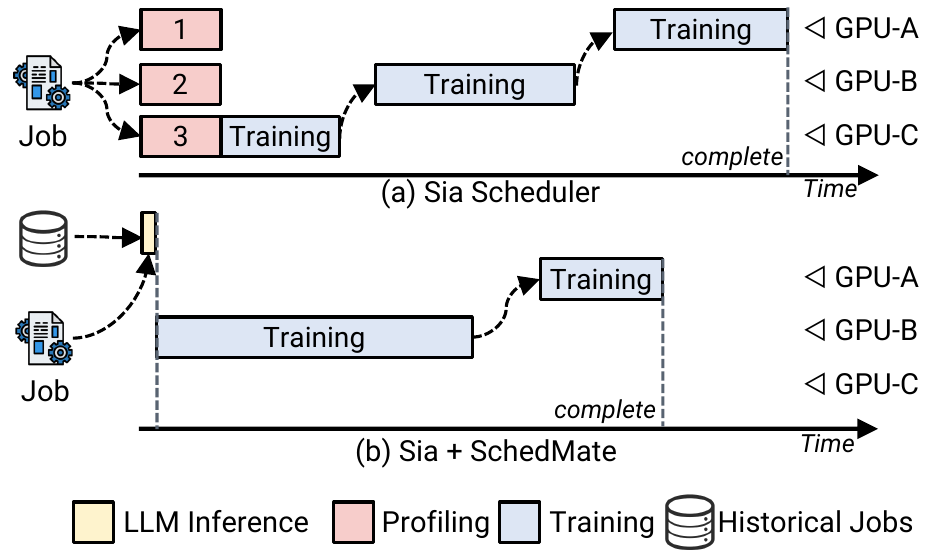}
    \caption{Job lifecycle in Sia and Sia+\SysName. The yellow block refers to the processing of \textit{Scheduling Advisor}. GPU-\{A,B,C\} represents different types of GPUs. }
\label{fig:sia_bypass_profiling}
\end{figure}

\noindent\textbf{Pre-profiling-free Bootstrapping.} We modify Sia's initial workflow. When a new job arrives, instead of immediately profiling it on GPUs, we first use the \textit{Scheduling Advisor} to retrieve similar historical jobs. If similar jobs are found, we reuse the historical performance model from the matched jobs to bootstrap the new job, allowing it to start training immediately without any profiling. If no match is found, we profile the job in the original way. Once the job is scheduled, we continue to apply Sia's profile-as-you-go method, updating the job's throughput information. Figure~\ref{fig:sia_bypass_profiling} illustrates this process. Vanilla Sia scheduler requires profiling on each GPU type before training, consuming additional resources. In contrast, our system leverages historical job data to bypass profiling, allowing training to start immediately and taking advantage of previous profiling results, which reduces overhead and improves cluster efficiency.

\subsection{Case Study 3: Standalone Scheduler \SysNameRM}
\label{subsec_integration_standalone}

We implement \SysName as a standalone scheduler following the vanilla Shortest Job First (SJF) policy. In this configuration, we employ both the \textit{Scheduling Advisor} and \textit{Failure Handler} modules. The standalone \SysName follows the vanilla Shortest Job First (SJF) policy. The scheduler maintains a database to store the scheduling metadata (submit time, end time, etc.) and workload fingerprint, extracted by Metadata Extractor, of historical jobs.

For each job, the \textit{Scheduling Advisor} retrieves three most similar historical jobs and calculates their average duration to estimate the new job's runtime. The scheduler then sorts the jobs based on these estimated durations and schedules them accordingly. In the event of a job failure, the \textit{Failure Handler} is triggered to analyze the root cause of the failure, take actions to mitigate the issue, and resume the job. This implementation allows us to evaluate \SysName's core capabilities in job duration estimation, efficient scheduling, and failure management independently of existing schedulers.

\cmmnt{\begin{table}[t]
    \caption{Case studies for \SysName integration. \textcolor{blue}{\ding{72}}: \textit{Scheduling Advisor}; \textcolor{red}{\ding{108}}: \textit{Metric Tracker}; \textcolor{green}{\ding{110}}: \textit{Failure Handler}.}
    \centering
    \resizebox{\linewidth}{!}{
    \small
    \setlength{\tabcolsep}{4pt}
    \begin{tabular}{@{}lcl@{}}
        \toprule
        \textbf{Scheduler} & \textbf{Mechanism} & \textbf{Key Benefits} \\
        \midrule
        Lucid \cite{Lucid} & Non-Intrusive & \begin{tabular}[c]{@{}l@{}}
            \textcolor{blue}{\ding{72}} Bypass profiling \\
            \textcolor{red}{\ding{108}} Interference-aware Evicting
        \end{tabular} \\
        \midrule
        Sia \cite{Sia} & Elastic & \textcolor{blue}{\ding{72}} Fast bootstrapping, Bypass profiling \\
        \midrule
        \SysName & Non-Intrusive & \begin{tabular}[c]{@{}l@{}}
            \textcolor{blue}{\ding{72}} Improve duration estimation \\
            \textcolor{green}{\ding{110}} Enable automatic failure recovery
        \end{tabular} \\
        \bottomrule
    \end{tabular}
    }
    \label{tab:case_studies}
\end{table}

% filepath: tables/baseline_policies.tex
\begin{table*}[t]
    \caption{Baseline scheduling policies compared against standalone \SysName (\S\ref{subsec_standalone}).}
    \centering
    \resizebox{\linewidth}{!}{%
    \small
    \setlength{\tabcolsep}{4pt} % Adjust column separation
    \begin{tabular}{@{}lcccc@{}}
        \toprule
        \textbf{Baseline} & \textbf{Scheduling Principle} & \textbf{Preemptive?} & \textbf{Intrusive?} & \textbf{Key Feature/Limitation} \\
        \midrule
        FIFO \cite{YARN, K8s} & Order of arrival & No & No & Simple, prone to Head-of-Line (HOL) blocking \\
        SJF & Shortest job first (Ideal) & No & No & Optimal avg. JCT (Impractical: requires future knowledge) \\
        QSSF \cite{Helios} & Predict shortest job (ML) & No & No & Data-driven prioritization, prediction-dependent \\
        Horus \cite{Horus} & Predict resource usage (ML) & No & Yes & Packing-enabled, requires model graph analysis \\
        Tiresias \cite{Tiresias} & Least attained service & Yes & Yes & Prioritizes short jobs implicitly, requires code changes \\
        \bottomrule
    \end{tabular}%
    }
    \label{tab:baseline_policies}
\end{table*}

\begin{table*}[htbp]
    \centering
    \caption{Comparison of Scheduling Policies Evaluated.}
    \label{tab:scheduling_policies_comparison}
    \resizebox{0.95\textwidth}{!}{%
    \begin{tabular}{l|l|l|l}
        \toprule
        \textbf{Scheduler} & \textbf{Profiling Method} & \textbf{Duration Estimation} & \textbf{Features} \\
        \midrule
        Lucid & Short profiling & ML model & Non-intrusive, Job packing \\
        Lucid+\SysName & \textit{Scheduling Advisor} &  Prediction & Interference-aware eviction \\
        \midrule
        Sia & Profile-as-you-go & Placement-aware profiling & Elastic scaling \\
        Sia+\SysName & Bypass profiling & Metadata-driven Prediction & Efficient bootstrapping \\
        \midrule
        Tiresias & ML-based & ML-based predictor & Preemptive scheduling \\
        QSSF & ML-based & ML-based short job prioritization & Shortest-job prioritization \\
        Horus & Historical job data & Historical analysis & Dependency-aware scheduling \\
        FIFO & - & - & First-in-First-out scheduling \\
        \midrule
        Standalone \SysName & \textit{Scheduling Advisor} & Metadata-driven Prediction & \textit{Failure Handler}+\textit{Scheduling Advisor} \\
        \bottomrule
    \end{tabular}%
    }
\end{table*}}

\section{Evaluation}
\label{sec_evaluation}

This section presents our evaluation of \SysName. We begin with experiments on a physical cluster (\S\ref{subsec_evaluation_physical}). We then assess \SysName's effectiveness in enhancing existing DL schedulers (\S\ref{subsec_lucid_case_study}, \S\ref{subsec_sia_case_study}) and as a standalone scheduler (\S\ref{subsec_standalone}).

\begin{figure}[t]
    \centering
    % Table content
    \captionof{table}{\textbf{Physical Evaluation}: Comparison of policy performance between physical and simulated environments.}
    \label{tab:physical_result}
    \resizebox{0.90\linewidth}{!}{\begin{tabular}{@{}llll@{}}
        \toprule
        \textbf{Policy}                                                                                                         & \textbf{Cluster} & \textbf{Avg. JCT} & \textbf{Avg. Queuing} \\ \midrule
    \multirow{2}{*}{Lucid+\SysName}                                                                                             & Physical         & \textbf{5.53h}             & \textbf{2.71h}             \\
                                                                                                                                & Simulation       & 5.64h             & 2.60h             \\ \midrule
        \multirow{2}{*}{Lucid}                                                                                                  & Physical         & 6.85h             & 3.40h             \\
                                                                                                                                & Simulation       & 6.79h             & 3.32h             \\ \bottomrule
        \end{tabular}}

    \vspace{1em}
    % \includegraphics[width=1.0\linewidth]{figures/simulation_fidelity.pdf}
    % \captionof{figure}{\textbf{Simulator Fidelity Validation.} JCT CDFs comparing simulation results (dashed lines) against physical cluster execution (solid lines) for (a) Lucid and (b) Lucid+\SysName.}
    % \label{fig:physical_result}
\end{figure}

\begin{table*}[t]
    \caption{DL model training workloads for scheduler evaluation on the physical cluster, categorized by scale.}
    \centering
    % \resizebox{0.95\textwidth}{!}{%
        \begin{tabular}{|p{3cm}|p{3cm}|p{2.5cm}|p{2.5cm}|p{3.5cm}|p{1.4cm}|}
            \hline
            \textbf{Scale} & \textbf{Model} & \textbf{Dataset} & \textbf{Batch Size} & \textbf{Task} & \textbf{\#Param.} \\
            \hline
            % Large Scale (>1B)
            \multirow{2}{*}{\textit{Large (>1B)}} & LLaMA-3~\cite{llama3} & Alpaca~\cite{alpaca}& 1$\sim$4 & Instruction Fine-tuning & 8B \\
            \cline{2-6}
             & Qwen-2.5~\cite{qwen2} & Alpaca~\cite{alpaca} & 1$\sim$4 & Instruction Fine-tuning & 14B \\
            \hline
            % Medium Scale (100M - 1B)
            \multirow{6}{*}{\textit{Medium (100M - 1B)}} & ViT-L/16~\cite{Vit} & ImageNet-1k~\cite{ImageNet} & 16$\sim$64 & Image Classification & 307M \\
            \cline{2-6}
             & BERT-Base~\cite{BERT} & SQuAD v1.1~\cite{SQuAD} & 16$\sim$64 & Fine-tuning & 110M \\
            \cline{2-6}
             & DeBERTa-Large~\cite{deberta} & SQuAD v1.1 & 16$\sim$64 & NLU Fine-tuning & 400M \\
            \cline{2-6}
             & Stable Diffusion~\cite{stablediffusion} & LAION~\cite{Laion-COCO} & 1$\sim$8 & Text-to-Image & 860M \\
            \cline{2-6}
             & CLIP ViT-L/14~\cite{clip} & WIT~\cite{wit} & 16$\sim$64 & Image Classification & 427M \\
            \cline{2-6}
             & DLRM~\cite{dlrm} & Criteo TB~\cite{criteo} & 512$\sim$2048 & Recommendation & $\sim$180M \\
            \hline
            % Small Scale (<100M)
            \multirow{7}{*}{\textit{Small (<100M)}} & ResNet50~\cite{ResNet} & ImageNet-1k~\cite{ImageNet} & 32$\sim$128 & Image Classification & 25.6M \\
            \cline{2-6}
             & Unet2D~\cite{Unet} & BraTS~\cite{brats}& 8$\sim$64 & Image Segmentation & 30M \\
            \cline{2-6}
             & EfficientNet-B4~\cite{EfficientNet} & ImageNet-1k~\cite{ImageNet} & 32$\sim$128 & Image Classification & 19M \\
            \cline{2-6}
             & ConvNeXt-Base~\cite{ConvNet} & ImageNet-1k~\cite{ImageNet} & 32$\sim$128 & Image Classification & 88M \\
            \cline{2-6}
             & MobileNetV2~\cite{mobilenetv2} & ImageNet-1k~\cite{ImageNet} & 32$\sim$128 & Image Classification & 3.5M \\
            \cline{2-6}
             & YOLOv8~\cite{yolov8} & COCO~\cite{coco} & 16$\sim$64 & Object Detection & 25.9M \\
            \cline{2-6}
             & T5-small~\cite{t5} & C4~\cite{t5} & 1$\sim$8 & Text Generation & 60M \\
            \hline
        \end{tabular}%
    % }
    \label{tab:physical_models}
\end{table*}

\subsection{Experiment Setup}

\noindent\textbf{Implementation.}~We implement \SysName on top of a Ray \cite{Ray}-based scheduling system with approximately 5,500 lines of code. \SysName is implemented as a library and can be integrated into existing schedulers. We deploy the core LLM as an API service using vLLM~\cite{vLLM}. For the embedding models, we use the FlagEmbedding library~\cite{bge2024chen}. We use Redis~\cite{Redis} for storing the vectors, historical job metadata, and performing similarity searching. The system incorporates real-time GPU monitoring using NVIDIA Management Library~\cite{nvml}, providing hardware metrics for profiling.

\noindent\textbf{Testbed and Models.}~We conduct our physical experiments on 8 nodes in a SLURM~\cite{SLURM} cluster. Each node is equipped with 8 NVIDIA A800-80GB GPUs and 2TB of memory. For software, we use Python 3.9, CUDA 12.2, PyTorch 2.4.0, RAY 2.6.3, and vLLM 0.5.5~\cite{vLLM}. In this section, if not specified, we use BGE-m3 \cite{bge2024chen} embedding model and Qwen-2.5-7B-Instruct \cite{qwen1.5} (enable GPTQ~\cite{GPTQ} Int8 quantization) as the core models. Additionally, we deploy the embedding model on CPU and LLM on a single GPU using vLLM. We compare the efficiency and accuracy of different models in \S\ref{subsec_microbenchmarks}.

\noindent\textbf{Traces.}~We use the following traces from various sources for comprehensive evaluation of \SysName:
\begin{itemize}[leftmargin=*,topsep=0pt, itemsep=-3pt]
    \item \textit{Mars}: A production trace we collect from a 1,024 GPU cluster, containing 500 LLM training jobs (7B-100B parameters) with code, logs, and hardware metrics. These jobs utilize the same codebase with about 40K LoC.
    \item \textit{Public Traces}: Widely-used public DL cluster traces including Acme~\cite{Acme}, Philly~\cite{Philly}, Helios~\cite{Helios} (Saturn and Venus), and Sia-trace~\cite{Sia}. Used for end-to-end physical and simulator evaluations.
    \item  \textit{Synthetic Traces}: We generated 100 different jobs from 20 open-source GitHub~\cite{GitHub} repositories covering diverse tasks and models in domains, such as CV, NLP, and audio. These repositories reflect varied development practices (e.g., code organization, logging, frameworks). \cmmnt{Table \ref{tab:opensource_repos} summarizes these repositories and their key properties.}
\end{itemize}

\cmmnt{\begin{table}[t]
    \caption{\textbf{Open-source repositories for synthetic traces\todo{need revision}} used in our evaluation. We select the repositories that are most representative of the real-world DL workloads.}
    \label{tab:opensource_repos}
    \centering
    \resizebox{0.95\linewidth}{!}{%
    \begin{tabular}{@{}lllll@{}}
    \toprule
    \textbf{Repository}          & \textbf{Framework} & \textbf{Task}        & \textbf{Dataset} \\ \midrule
    LLaMA3 8B ~\cite{llama3} & Transformer        & Finetuning           & Alpaca \\
    Unet2D~\cite{Unet}       & DeepSpeed          & Image     & Alpaca  \\
    ResNet50~\cite{ResNet}   & Detectron2          & Image & Laion   \\
    ViT~\cite{Vit}           & InternEvo          & Image & COCO    \\
    BERT~\cite{BERT}         & MegatronLM         & Pretraining          & Github  \\ \bottomrule
    \end{tabular}}
\end{table}
}

\noindent\textbf{Workloads.}~In the physical experiments, we use the models, datasets, and batch sizes in Table \ref{tab:physical_models}. In the simulation experiments of standalone \SysName (\S\ref{subsec_standalone}), we use a sampled one-month trace from Acme without changing the workload, mainly LLM training jobs. In the Lucid (\S\ref{subsec_lucid_case_study}) and Sia(\S\ref{subsec_sia_case_study}) simulation experiments, we follow the workload recipes of the original papers. To simulate real-world development practices, we set a portion of the job to exit early.

% \input{tables/model_speed.tex}

% \noindent\textbf{Baselines.}~We compare \SysName against state-of-the-art DL schedulers, and other recently proposed DL schedulers (Themis~\cite{Themis}, Optimus~\cite{optimus}, Horus~\cite{Horus}, QSSF~\cite{Helios}, Tiresias~\cite{Tiresias}).

% \todo{Maybe remove the following part. Add some explanation about the policies (table).}
% \begin{itemize}[leftmargin=*,topsep=0pt, itemsep=0pt]
%     \item \textit{Physical Cluster Evaluation} (\S\ref{subsec_evaluation_physical}): We assess the real-world performance gains of Lucid+\SysName and validate the fidelity of our simulation result on a physical cluster with 64 NVIDIA A800 GPUs.
%     \item \textit{Simulation-based Integration Studies} (\S\ref{subsec_evaluation_simulation}): We evaluate \SysName's integration with state-of-the-art schedulers through three case studies: (1) \textit{Non-intrusive scheduling with Lucid}, where we quantify improvements in job completion times and resource utilization through enhanced profiling and our novel interference-aware eviction mechanism; (2) \textit{Elastic scheduling with Sia}, where we demonstrate how our scheduling advisor enhances scheduling decisions in heterogeneous GPU environments; and (3) \textit{Standalone scheduling}, where we deploy \SysName as an independent scheduler.
%     \item \textit{Module-specific Evaluation} (\S\ref{subsec_modules_evaluation}): We evaluate the individual components of \SysName. We also analyze the system's overhead, scalability, and performance across different LLM models.
% \end{itemize}

\begin{figure*}[t]
    \centering
    % \vspace{-8pt}
    \includegraphics[width=0.95\linewidth]{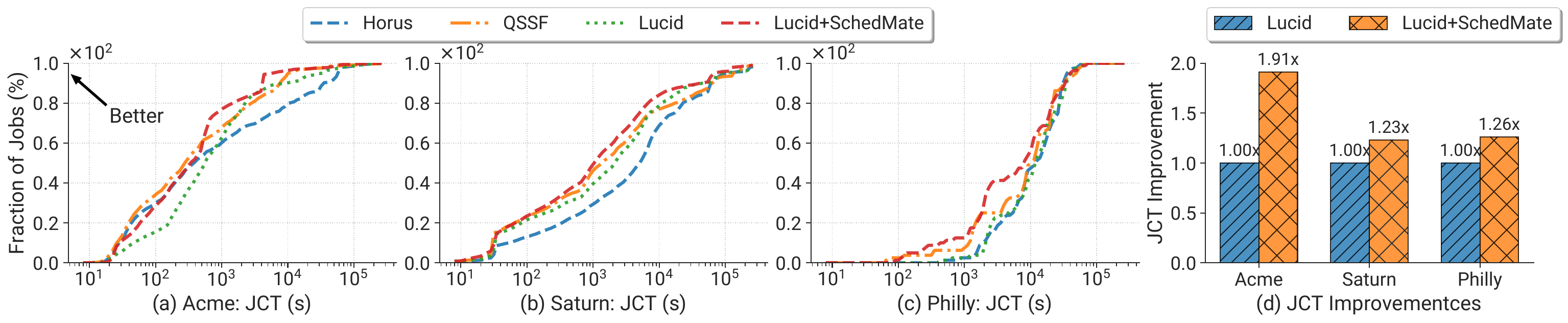}
    \caption{\textbf{Case Study 1.} The CDF curves of JCTs using different scheduling policies (a,b,c) and (d) Avg. JCT improvement of Lucid+\SysName against Lucid on four traces: Acme, Philly, Saturn, and Venus.}
\label{fig:bar_lucid_jct}
\end{figure*}

\subsection{Evaluation on a Physical Cluster}\label{subsec_evaluation_physical}

We evaluate Lucid and Lucid+\SysName (settings in \S\ref{subsec_integration_nonintrusive}) on a physical cluster comprising 8 nodes, each equipped with 8 NVIDIA A800-80GB GPUs (128 GPUs total). Our implementation extends the RAY~\cite{Ray} scheduler to incorporate Lucid and \SysName policies, utilizing RAY's fractional GPU capabilities for job packing. Each job is submitted as a RAY task. This approach allows deployment on existing clusters without altering the underlying resource manager.

The evaluation workload consists of 241 jobs sampled from Philly and Acme. We assign workloads from Table \ref{tab:physical_models} based on the job's original GPU request size: \textit{large} (8 GPUs), \textit{medium} (4-8 GPUs), and \textit{small} (<4 GPUs). Job requests exceeding the cluster capacity are capped at 128 GPUs. The slowdown threshold is set to 0.5. We repeat 3 times to eliminate randomness. Lucid's profiler reserves one node and profiles each job for 100s. \SysName operates as a RAY actor on a dedicated GPU, where we deploy Qwen-2.5-7B-Instruct and BGE-m3 as core models. Results in Table \ref{tab:physical_result} show that Lucid+\SysName reduces the average JCT by 23.2\% compared to Lucid on the physical cluster. This improvement comes from \SysName's ability to cancel subpar packing decisions and enhanced duration estimation. \cmmnt{Specifically, we observe underestimation by Lucid's profiler, leading to suboptimal job packing decisions.} We also observe that the average latency of \textit{Scheduling Advisor} is 9s, which is acceptable compared to the profiling overhead.

We also verify the fidelity of our simulator. We use the simulator in \ref{subsec_lucid_case_study} to process the same traces and settings and compare the result with the ground truth result. The average error rate of both average JCT and makespan is less than 3.7\%, which indicates the high fidelity of our simulator.

% \begin{figure}[t]
%     \centering
%     \includegraphics[width=0.9\linewidth]{figures/case_lucid_cdf.pdf}
%     \caption{CDF curves of JCTs for Lucid and Lucid+\SysName across four clusters. X-axis of each subfigure is in log scale.}
%     \label{fig:cdf_lucid_case_study}
% \end{figure}

\subsection{Simulation-based Evaluation}\label{subsec_evaluation_simulation}

% \noindent\textbf{Simulator Settigns.}~

\noindent\textbf{Case Study 1: Non-intrusive scheduling with Lucid.}\label{subsec_lucid_case_study}~We conduct an end-to-end simulation on the performance of Lucid+\SysName, Lucid, and Horus, QSSF, on four traces: Acme, Philly, Saturn, and Venus.
Figure~\ref{fig:bar_lucid_jct}(a,b,c) presents the CDF curves of JCTs on different traces. Lucid+\SysName consistently outperforms other policies. Figure~\ref{fig:bar_lucid_jct}(d) presents the average JCT improvements of Lucid+\SysName against Lucid. Lucid+\SysName demonstrates superior performance, improving $1.23\times\sim1.91\times$ compared to Lucid alone.
The key source of improvements is its ability to dynamically detect slowdowns of job packing and predict workload using \textit{Workload Metadata}. Specifically, among the traces, the performance gain of \SysName is more significant on the Acme trace, owing to the fact that Acme trace contains heavier workloads, which leads to a significant slowdown in job packing. The interference-aware eviction policy of \SysName is more effective in such traces. Besides, Acme has a high-skewed workload distribution~\cite{Acme}, where most of the jobs are short-term jobs, introducing challenges to duration estimation. In addition, the improvements in other traces are smaller because the duration distribution is more concentrated in Saturn and Philly, and the jobs are relatively lightweight. Many short jobs are finished during profiling. The interference of jobs is also rather lower, leaving less room for the improvement of the eviction policy.

\begin{table}
    \caption{\textbf{Case Study 2.} Comparison of the performance of Sia and Sia+\SysName across different traces.}    \centering
    \resizebox{0.9\linewidth}{!}{
    \begin{tabular}{@{}lllll@{}}
        \toprule
        \textbf{Trace} & \textbf{Policy}                                     & \textbf{Avg. JCT}                                        & \textbf{p99 JCT}                                              & \textbf{Makespan}                           \\ \midrule
                                   & Sia                                                          & 0.74h                                                & 10.60h                                             & 14.7h                                                \\
        \multirow{-2}{*}{Philly}   & \textbf{Sia+\SysName}                                    & \textbf{0.59h}                                       & \textbf{10.12h}                                    & \textbf{14.4h}                                       \\ \midrule
                                   & Sia                                                          & 0.83h                                      & 12.0h                                              & 15.7h                                                \\
        \multirow{-2}{*}{Helios}   & \textbf{Sia+\SysName}                                    & \textbf{0.72h}                                       & \textbf{10.7h}                                     & \textbf{14.8h}                                       \\ \midrule
                                   & Sia                                                          & 0.64h                                                & 4.25h                                               & 12.5h                                                \\
        \multirow{-2}{*}{Sia-trace} & \textbf{Sia+\SysName}                                    & \textbf{0.53h}                                       & \textbf{4.12h}                                     & \textbf{12.4h}                                       \\ \bottomrule
    \end{tabular}
    }
    \label{sia_case_study}
\end{table}

\noindent\textbf{Case Study 2: Elastic Scheduling with Sia.}\label{subsec_sia_case_study}~As mentioned in \S\ref{subsec_integration_elastic}, Sia+\SysName can reduce the profiling overhead through \textit{Preprofiling-free Bootstraping}. To test the effectiveness of \SysName in elastic scheduling scenarios, we integrate \SysName into Sia's simulator and conduct evaluation using identical workload and heterogeneous cluster settings in Sia's paper~\cite{Sia}. Table \ref{sia_case_study} presents the results. We observe that our system consistently outperforms Sia. Specifically, \SysName reduces the average JCT by $13.3\%\sim20\%$. Furthermore, the p99 JCT and makespans also show consistent improvements. The improvements derive from the benefits of \textit{Scheduling Advisor} in mitigating profiling overhead while maintaining the accuracy of the scheduling decisions. In summary, this case study demonstrates that integrating \SysName with Sia (\textbf{Sia+\SysName}) significantly enhances scheduling efficiency.

\begin{figure}[t]
    \centering
    \includegraphics[width=1.0\linewidth]{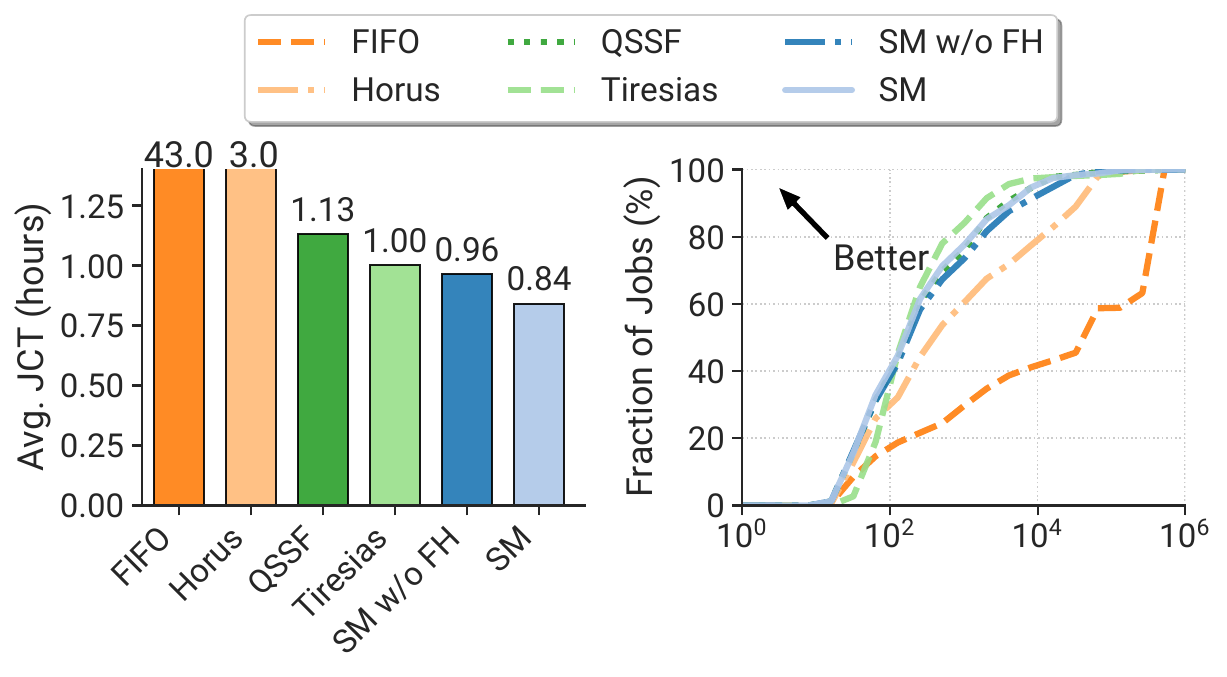}
    \caption{\textbf{Case Study 3: Standalone \SysName Evaluation.} (Left) Job Completion Times (JCTs) and queuing delays. (Right) CDF curves of JCTs using different policies. \textbf{SM}: \SysName. \textbf{FH}: \textit{Failure Handler}.}    \label{fig:standalone_bar_cdf}
\end{figure}

\noindent\textbf{Case Study 3: Standalone Scheduling.}\label{subsec_standalone}~We evaluated the performance of \SysName as a standalone scheduler, which utilizes \textit{Scheduling Advisor} for duration estimation and \textit{Failure Handler} for automatic failure recovery. We sample jobs of Acme between June and August 2023. Specifically, to evaluate the failure recovery performance, we identify resumed failed jobs from the trace and consider failures in the simulator, which is the first simulator considering this factor. We compared standalone \SysName against several baseline schedulers: Tiresias~\cite{Tiresias} (a preemptive scheduler), Quasi-Shortest-Service-First (QSSF)~\cite{Helios} (ML to prioritize short jobs), FIFO, and Horus~\cite{Horus}. Figure~\ref{fig:standalone_bar_cdf} shows that standalone \SysName achieves a 25.7\% improvement in average JCT over QSSF and 16\% over Tiresias, primarily due to its superior duration estimation capabilities provided by the \textit{Scheduling Advisor}. Furthermore, incorporating the \textit{Failure Handler} yields an additional $\sim$12.5\% reduction in average JCT compared to \SysName without failure handling, highlighting its effectiveness in mitigating the impact of job failures. Overall, this case study demonstrates that standalone \SysName significantly enhances scheduling efficiency through LLM-based workload prediction and failure management.

\subsection{Module-Specific Performance Analysis}\label{subsec_modules_evaluation}

\begin{figure}[t]
    \centering
    \includegraphics[width=1.0\linewidth]{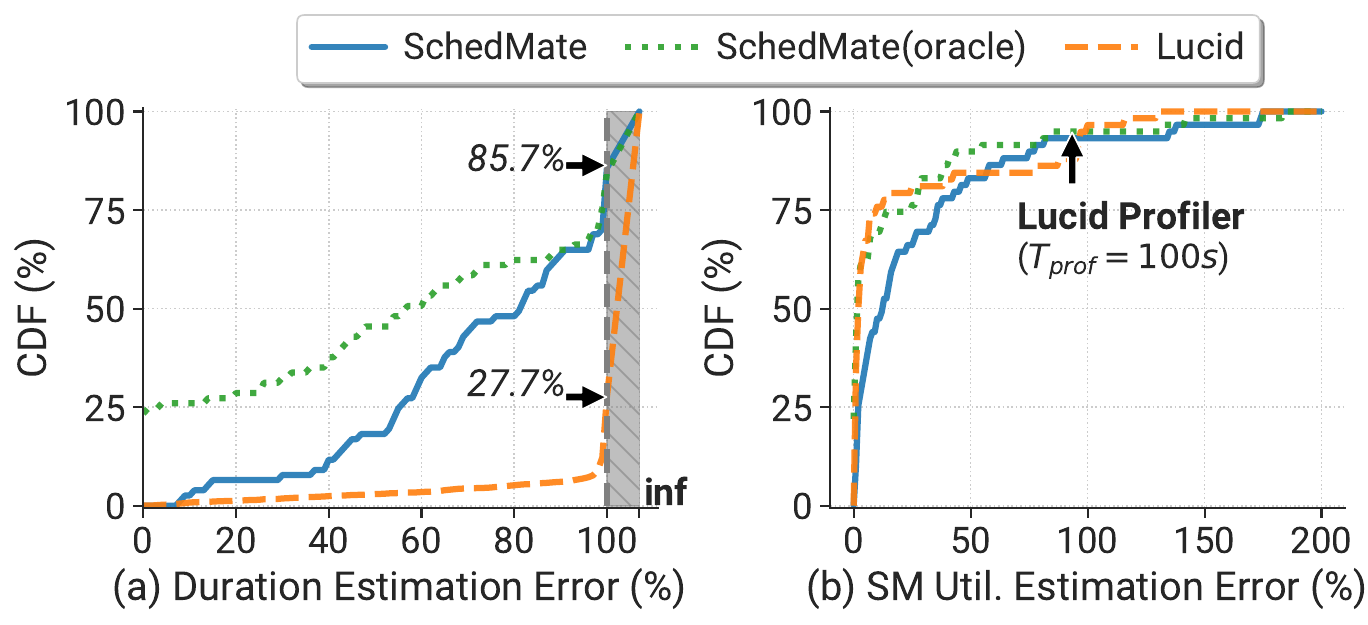}
    \caption{CDF curves of relative error for (a) duration estimation and (b) SM utilization estimation of \SysName, oracle \SysName, and Lucid.}
    \label{fig:duration_estimation_error_lucid}
\end{figure}

%We conduct an experiment to demonstrate the performance of \textit{Scheduling Advisor}. Specifically,
\noindent\textbf{Scheduling Advisor Evaluation.}~We compare the performance of workload estimation of \textit{Scheduling Advisor} against an ML-based estimator from Lucid. We use the Mars trace to evaluate both methods. Specifically, the Lucid estimator trains an $GA^2M$ model~\cite{Lucid} for duration estimation.
Figure \ref{fig:duration_estimation_error_lucid} illustrates the results across \SysName, Lucid, and oracle \SysName, which always gets the correct similar jobs from historical jobs. Due to the sensitivity of relative error when ground truth values are small, we truncate errors in (a) and (b). For duration estimation, \SysName achieves relative errors less than 100\% for 85.7\% of estimations, significantly outperforming Lucid at 27.7\%. Notably, the curve shows a sharp increase as the relative error approaches 100\%. Regarding SM utilization estimation, \SysName performs slightly below Lucid's profiler, with average relative errors of 28.1\% and 19.3\%, respectively. However, \SysName's method requires less than 10 seconds on a single GPU per job, whereas Lucid demands profiling per job for a period of time ($T_{prof}=100s$ in this case). Furthermore, oracle \SysName presents a promising performance in both tasks, demonstrating the potential of our retrieval-based method.

% \input{tables/eval_metric_tracker.tex}

% \todo{Change to Qwen2.5-14B}One interesting observation is that models with sizes of 7B/8B like LLaMA 3.1-8B and Qwen 2.5-7B struggle to follow the instruction in Figure~\ref{figure_llm_prompt} for metadata extraction task. We observe that these models typically process only a limited number of files and terminates early, resulting in incomplete answers. In contrast, Qwen1.5-14B can perform the task effectively while maintaining relatively high efficiency.

% \ana{Yes, make sure to report the compute time and cost spent to extract this information and compare it to the overall compute time and cost benefits achieved due to enhanced cluster scheduling policy decisions. ie., do you win overall? is this extra effort really worth it in the end? Is the LLM extraction overhead something you just pay once and then you plug in the result to the scheduler and use it over and over again, amortizing the cost of the LLM extraction? Or is it an ongoing cost/overhead you are always incurring as the DL scheduler runs? }

% \zerui{Yes, overhead would be an important concern for employing LLMs. I have moved the evaluation of overhead and each component to the micro-benchmark section. I will finalize the remaining ones and update this section as soon as possible.}

\begin{table}[htbp]
    \centering
    \caption{\textbf{Failure Handler Evaluation}. We evaluate the component in identifying infrastructure failures against baselines. Failure Handler utilizes Qwen-2.5 models.}
    \resizebox{0.8\linewidth}{!}{%
    \begin{tabular}{@{}lcccc@{}}
        \toprule
        \multicolumn{1}{c}{\textbf{Method}}                                                  & \textbf{Model} & \underline{\textbf{F1-score}} & \textbf{Precision} & \textbf{Accuracy} \\ \midrule
        \multirow{3}{*}{Failure Parser}                                                      & 7B             & \textbf{67.7}    & \textbf{75.7}               & \textbf{90.1}              \\
                                                                                             & 14B            & \textbf{65.1}    & \textbf{77.1}               & \textbf{90.0}              \\
                                                                                             & 32B            & \underline{68.2}    & \underline{78.4}               & \underline{90.7}              \\ \midrule
        \multirow{3}{*}{\begin{tabular}[c]{@{}l@{}}Failure Parser\\ w/o Locator\end{tabular}} & 7B             & 11.1                & 75.0               & 83.8              \\
                                                                                             & 14B            & 62.5                & 47.9               & 81.8              \\
                                                                                             & 32B            & \textbf{69.0}       & \textbf{81.1}      & \textbf{90.9}     \\ \midrule
        RCACopilot~\cite{RCACopilot}                                                                           & FastText       & 43.0                & 50.0               & 87.0              \\ \bottomrule
        \end{tabular}
    }
    \label{tab:eval_failure_handler}
\end{table}

\noindent\textbf{Failure Handler Evaluation.}~We evaluate our failure handler on 300 failed jobs sampled from Mars, among which 75 are infrastructure failures. We set \textit{failure handler w/o locator} as the baseline, which simply parses the \underline{last 500} log lines without attempting to locate the failure message, relying on the assumption that the relevant information is likely to be near the end of the log. For the \textit{failure handler}, we parse the \underline{200} log lines around the located failure message. Results are summarized in Table \ref{tab:eval_failure_handler}. We observe that the \textit{Failure Handler} method consistently outperforms or matches the baseline across different model sizes, suggesting reliable identification of infrastructure failures with minimal false positives. While the baseline only shows competitive performance with the 32B model, our method achieves comparable results with smaller models, indicating better efficiency. Specifically, the 7B model achieves a comparable F1-score against the 32B model. This demonstrates that the failure locator effectively narrows down the relevant log lines, enabling smaller models to perform well without sacrificing accuracy, thereby reducing both latency and resource consumption.

We also compare our system against RCACopilot~\cite{RCACopilot}, a recent embedding-based approach. It achieves only 43.0 F1-score. We believe this is due to the complex failure patterns, which are not well captured by the embedding model.

%Notably, infrastructure failures only account for 16.7\% in the dataset, so the accuracy may not be a good metric in this case.

\begin{figure}[t]
    \centering
    \includegraphics[width=\linewidth]{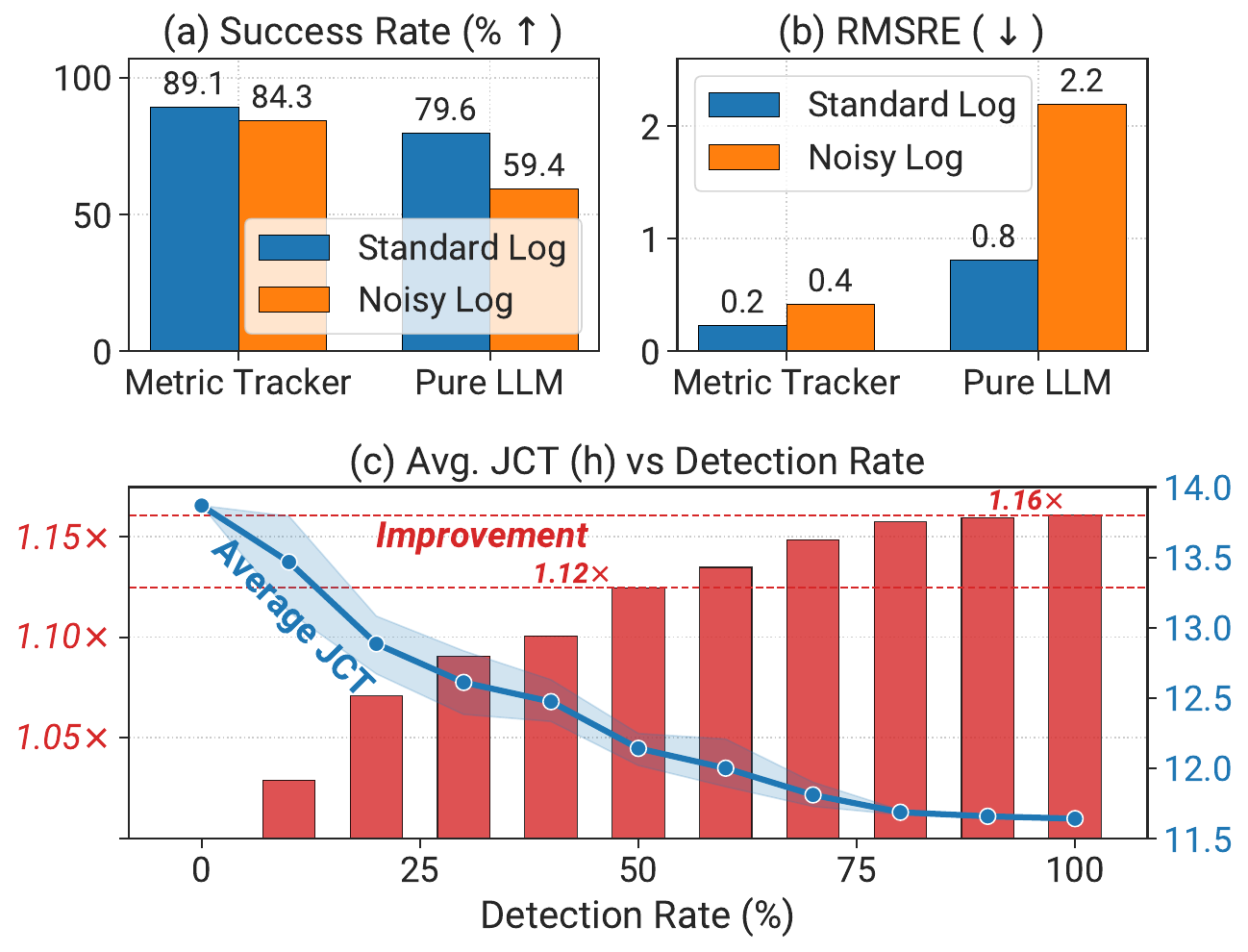}
    \caption{\textbf{Metric Tracker Evaluation.} Performance comparison between \textit{Metric Tracker} and a pure LLM under standard and noisy conditions. (a) Success rate evaluation. (b) Rmsre comparision. (c) Impact of slowdown detection rate on average JCT (red) and its improvement (blue) in Lucid+\SysName (each repeated five times).}
    \label{fig:eval_metric_tracker}
\end{figure}

\noindent\textbf{Metric Tracker Evaluation.}~The \textit{Metric Tracker} resolves the noisy log challenges in real-world scenarios. We tested it on a synthetic dataset of 1,000 logs, with 50\% containing massive irrelevant data. We set the pure LLM method as the baseline, which processes the log without filtering. We perform \textit{step time extraction} using both methods. As shown in Figure \ref{fig:eval_metric_tracker}, \textit{Metric Tracker} achieved an 84.3\% success rate and an RMSRE of 0.42, significantly outperforming the pure LLM approach, whose success rate drops to 59.2\%. These results highlight the \textit{Metric Tracker}'s superior resilience to noisy log data. Besides, we also observe that our method has a lower latency with \textbf{8.7}s per log compared to 9.8s per log in pure LLM. These results demonstrate both the efficiency of the \textit{Metric Tracker} and robustness in handling noisy data, which is common in real-world DL job logs.

\subsection{Micro-benchmarks}
\label{subsec_microbenchmarks}

% Robustness of metric tracker & failure handler
\noindent\textbf{Ablation Study on Partial Information.}~The non-intrusiveness of \SysName indicates the absence of certain information. We perform an ablation study on the Lucid+\SysName's packing eviction technique. In practice, we can only detect the slowdown of jobs when they actually output the step times in their logs. To simulate different levels of partial information, we manually control the \textbf{detection rate} from 0 to 1.0. The detection rate refers to the percentage of detected slowdown jobs.  As shown in Figure~\ref{fig:eval_metric_tracker}, the JCTs of Lucid on a sampled \textbf{Acme} trace with interference-aware evicting under different slowdown detection rates are presented. We assign the workloads in Table \ref{tab:physical_models} and set the slowdown threshold to 0.50. The JCTs under different settings are presented in Figure~\ref{fig:eval_metric_tracker}(c). We observe that the average JCT decreases as the detection rate increases. The improvement achieves 12.5\% when the detection rate is 0.5. When the detection rate is above 0.6, the performance gain is limited. \textit{Metric Tracker} relies on the assumption that most jobs would output the metrics in the log. In addition, the result in Figure \ref{fig:eval_metric_tracker} shows \textit{Metric Tracker} achieves a 91.6\% success rate using a Qwen-2.5-7B model. \cmmnt{This suggests that if about 60\% of the jobs record the metrics in their logs, we can already achieve considerable performance gain with \textit{Metric Tracker}.} In summary, relying on users' logs to detect the slowdown of jobs seems to be a risky approach. However, we prove that in real-world scenarios, partial information can achieve considerable performance gain.

\begin{figure}[t]
    \centering
    \includegraphics[width=0.90\linewidth]{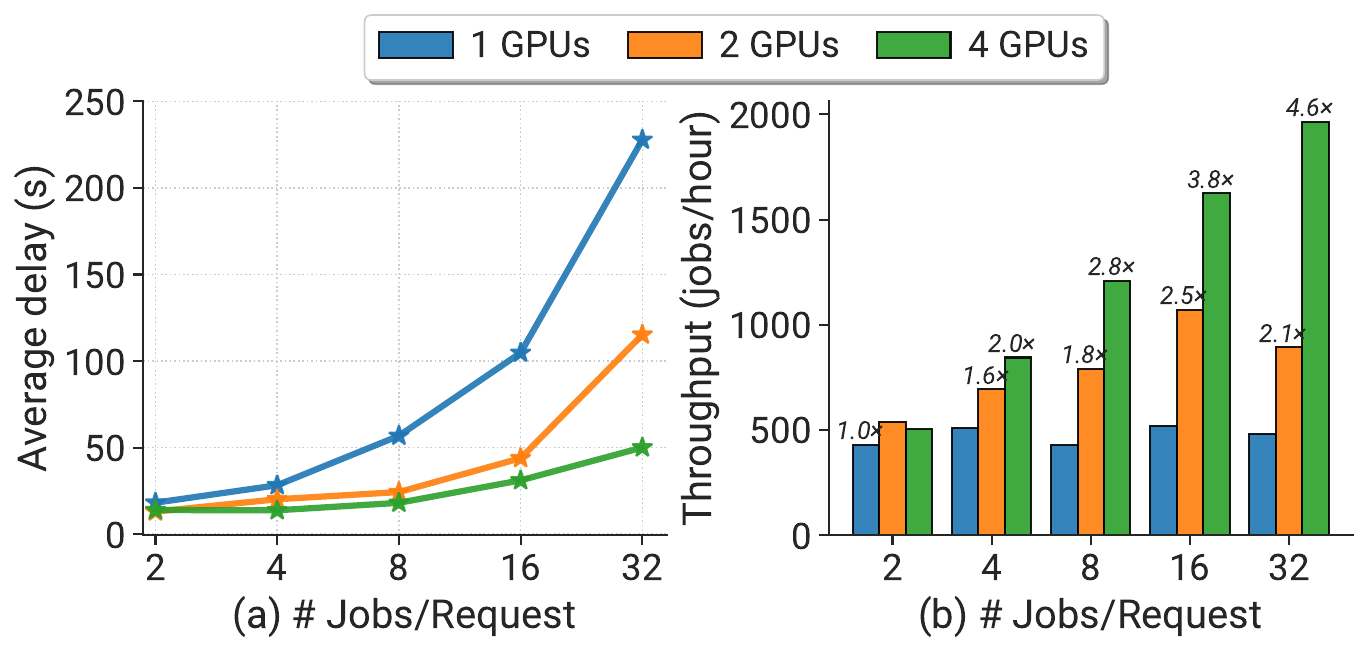}
    \caption{Performance scaling with job intensity and GPU count of the \textit{Scheduling Advisor} using a Qwen-1.5-14B-Int8 model (a) Average delay vs. jobs per request. (b) Throughput and improvement ratios for different numbers of GPUs.}
    \label{fig:micro_scalability}
\end{figure}

\begin{figure}
    \centering
    \includegraphics[width=0.8\linewidth]{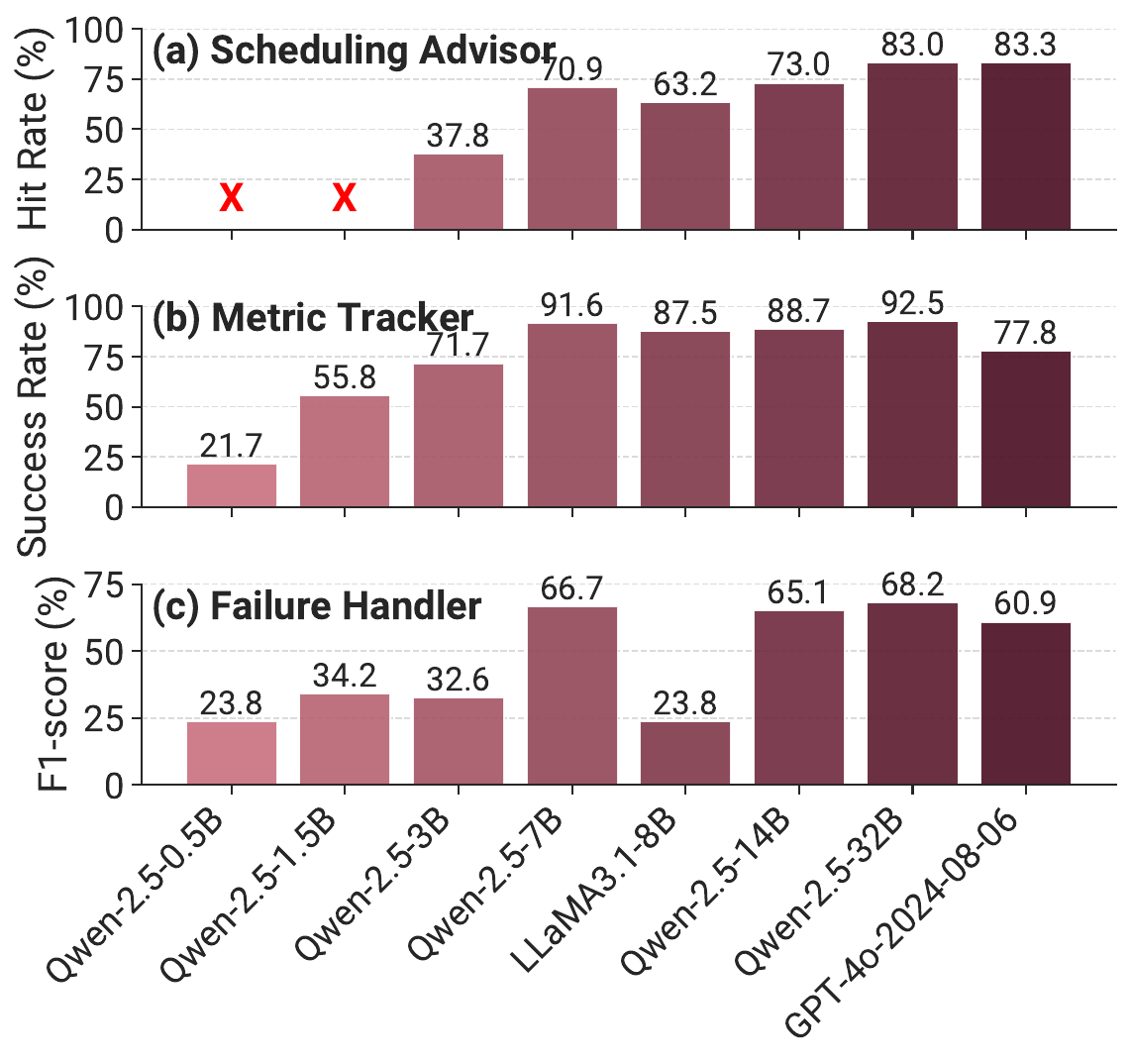}
    \caption{\textbf{Evaluation on different LLMs.} We involve 7 open-source models and one closed-source model, GPT-4o. The red cross mark means that the model fails in the task.}
    \label{fig:different_llms}
\end{figure}

\noindent\textbf{Impact of Different LLMs.}~Figure \ref{fig:different_llms} presents the performance of various LLMs across different scales for the three modules in \SysName. We observe that model performance improves with increased model size, with Qwen2.5-32B achieving the best overall results among open-source models. Smaller models (0.5B and 1.5B) failed to perform the SA task, indicating that a certain level of model complexity is required for effective metadata extraction. Interestingly, the closed-source GPT-4o model demonstrates superior performance in the SA task but unexpectedly underperforms in MT and FH tasks compared to larger Qwen models. In addition, despite competitive performance in MT tasks, LLaMA-3.1-8B struggles significantly in the FH task and does not match the same scale Qwen models in the other two tasks. The results highlight a trade-off between model size, task performance, and computational requirements, with models in the 7B to 14B parameter range offering a good balance of effectiveness and efficiency for \SysName's tasks.

\noindent\textbf{Scalability and Overhead}\label{subsec_scalability}~We sample 200 jobs from \textit{Mars} and test the \textit{Scheduling Advisor}, which takes the heaviest load among the three modules, under varying GPU counts and request intensity. We deploy Qwen-1.5-14B-Int8 model on each GPU. Note that for testing scalability on a heavier workload, we only sample large-scale jobs that use the same project with over 40K LoC.

% \footnote{Due to resource constraints, we only test on this version, which has slightly better performance and $\sim0.8\times$ throughput against Qwen-2.5-7B.}

Figure \ref{fig:micro_scalability} (a) shows that when the request intensity is less than 4, the Avg. delay remains less than 20s, acceptable for one GPU and no significant benefit for more. When request intensity $\geq8$, delay of one GPU increases significantly, while using more GPUs maintains lower delays. Figure \ref{fig:micro_scalability} (b) illustrates the throughput improvements. We set the \textit{1GPU2job throughput} as the baseline and mark the throughput improvement. It's obvious that single GPU throughput remains steady ($\sim$480 jobs/hour), while the 2-GPU and 4-GPU settings increase. Specifically, the 2-GPU setting achieves over 2$\times$ throughput when request intensity reaches 16, and 4-GPU up to 4.6$\times$ ($\sim$2000 jobs/hour) throughput over the baseline setting when 32 jobs per request. In summary, we also recommend large-scale cluster operators to dynamically assign GPUs according to the actual submission rate. For instance, for PAI~\cite{MLaaS} with periodical load ($300\sim1200$ jobs/h), while a single GPU is suitable for low load periods, to manage peak load periods, one should allocate 4 or more GPUs.

%-------------------------------------------------------------------------------

\section{Discussion and Related Work}
\label{sec_discussion}
%-------------------------------------------------------------------------------

\noindent\textbf{Supporting Other Workloads.}~While \SysName targets DL training, its core principle of semantic-aware scheduling is applicable to other domains like big data analytics~\cite{GoogleTrace12} and serverless computing~\cite{ServerlessATC20}. The \textit{Scheduling Advisor}, for instance, could be adapted to extract domain-specific semantics to address challenges such as data locality for big data jobs or dynamic scaling for serverless functions. Future work could explore these adaptations and integrate more advanced LLM agent techniques~\cite{AgentSurvey} to more intelligently extract and utilize semantic information.

\noindent\textbf{Related Work.}~Existing DL schedulers~\cite{Pollux,Tiresias,Sia,Gavel,Lucid,AutoSched,Helios,Acme,AntMan,Horus,Prism,ONES} make efforts to optimize resource allocation and job scheduling but works without enough semantic information of jobs. Machine learning has been increasingly used to optimize systems~\cite{Lucid,Primo,NetLLM,Sibylla}. More recently, LLMs have been applied to specific system tasks like log parsing~\cite{LogPPT,LLMParser} and failure diagnosis~\cite{RCAgent,RCACopilot,Ciri}. However, no works have been proposed to extract deep semantic information of cluster jobs to enhance DL job scheduling decisions. \SysName is distinct in its holistic approach: it harnesses LLMs to bridge the semantic gap between DL jobs and the cluster scheduler, enabling a new paradigm of semantic-aware DL cluster scheduling.

\section{Conclusion}

\SysName introduces a new paradigm of semantic-aware scheduling that extracts semantic insights of jobs using LLMs. Our evaluations show that it enhances state-of-the-art DL schedulers by improving workload prediction, enabling dynamic runtime intervention, and automating failure recovery to significantly boost cluster performance and efficiency.

\bibliographystyle{plain}
\bibliography{ref}

\end{document}